\documentclass[USenglish,oneside,twocolumn]{article}

\usepackage[utf8]{inputenc}
\usepackage[big]{dgruyter_NEW}
 
\DOI{foobar}

\cclogo{\includegraphics{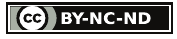}}

\usepackage{amsfonts}       

\usepackage{algorithm}
\usepackage{algorithmic}
\makeatletter
\def\plist@algorithm{\ALG@name\space}
\makeatother

\usepackage{graphicx}
\usepackage{subcaption}

\usepackage{pgfplots}
\usepackage{filecontents}
\usepackage{amsmath}
\usepackage{amsthm}
\usepackage{stmaryrd}

\newcommand{\descr}[1]{\vspace{0.2cm} \noindent \textbf{#1}}

\newcommand{\bc}[1]{\left\{{#1}\right\}}
\newcommand{\br}[1]{\left({#1}\right)}
\newcommand{\bs}[1]{\left[{#1}\right]}
\newcommand{\abs}[1]{\left| {#1} \right|}

\newcommand{\ip}[2]{\left\langle{#1},{#2}\right\rangle}
\newcommand{\norm}[1]{\left\| {#1} \right\|}

\renewcommand{\P}[1]{\mathbb{P}\bs{{#1}}}

\newcommand{\Ee}[2]{\underset{#1}{\mathbb{E}}\bs{{#2}}}

\newtheorem{theorem}{Theorem}
\newtheorem{proposition}{Proposition}
\newtheorem{fact}{Fact}

\newtheorem{definition}{Definition}
\theoremstyle{definition}
\newtheorem{example}{Example}

\newcommand{\ind}[1]{\llbracket {#1} \rrbracket}

\makeatletter
\newcommand{\newreptheorem}[2]{\newtheorem*{rep@#1}{\rep@title} 
	\newenvironment{rep#1}[1]{\def\rep@title{#2 \ref*{##1}}\begin{rep@#1}}{\end{rep@#1}}
}
\makeatother

\newreptheorem{lemma}{Lemma}
\newreptheorem{theorem}{Theorem}
\newreptheorem{claim}{Claim}
\newreptheorem{proposition}{Proposition}
\newreptheorem{corollary}{Corollary}

\allowdisplaybreaks

\begin{document}

  \author*[1]{Parameswaran Kamalaruban}

  \author[2]{Victor Perrier}

  \author[3]{Hassan Jameel Asghar}

  \author[4]{Mohamed Ali Kaafar}

  \affil[1]{\'Ecole Polytechnique F\'ed\'erale de Lausanne (work done while Kamalaruban was a Postgraduate Researcher at Data61, CSIRO), E-mail: kamalaruban.parameswaran@epfl.ch}

  \affil[2]{ISAE-SUPAERO \& Data61, CSIRO, E-mail: v.perrier0@gmail.com}

  \affil[3]{Macquarie University \& Data61, CSIRO, E-mail: hassan.asghar@mq.edu.au}

  \affil[4]{Macquarie University \& Data61, CSIRO, E-mail: dali.kaafar@mq.edu.au}
  
%
%
%
%

\title{\huge Not All Attributes are Created Equal: $d_{\mathcal{X}}$-Private Mechanisms for Linear Queries}

\runningtitle{Not All Attributes are Created Equal: $d_{\mathcal{X}}$-Private Mechanisms for Linear Queries}


\begin{abstract}
{Differential privacy provides strong privacy guarantees simultaneously enabling useful insights from sensitive datasets. However, it provides the same level of protection for all elements (individuals and attributes) in the data. There are practical scenarios where some data attributes need more/less protection than others. In this paper, we consider $d_{\mathcal{X}}$-privacy, an instantiation of the privacy notion introduced in \cite{chatzikokolakis2013broadening}, which allows this flexibility by specifying a separate privacy budget for each pair of elements in the data domain. We describe a systematic procedure to tailor any existing differentially private mechanism that assumes a query set and a sensitivity vector as input into its $d_{\mathcal{X}}$-private variant, specifically focusing on linear queries. Our proposed meta procedure has broad applications as linear queries form the basis of a range of data analysis and machine learning algorithms, and the ability to define a more flexible privacy budget across the data domain results in improved privacy/utility tradeoff in these applications. We propose several $d_{\mathcal{X}}$-private mechanisms, and provide theoretical guarantees on the trade-off between utility and privacy. We also experimentally demonstrate the effectiveness of our procedure, by evaluating our proposed $d_{\mathcal{X}}$-private Laplace mechanism on both synthetic and real datasets using a set of randomly generated linear queries.} 		
\end{abstract}


\journalname{Proceedings on Privacy Enhancing Technologies}
\DOI{Editor to enter DOI}
\startpage{1}
\received{..}
\revised{..}
\accepted{..}

\journalyear{..}
\journalvolume{..}
\journalissue{..}

\maketitle

\section{Introduction}
Differential privacy~\cite{dwork2006calibrating} is a formal notion of privacy that allows a trustworthy data curator, in possession of sensitive data from several individuals, to approximately answer a set of queries submitted by an analyst while maintaining individual privacy. 
Intuitively, differential privacy guarantees that query answers with or without any individual's data are (almost) indistinguishable. One common mechanism for achieving differential privacy is to inject random noise to the query answers, carefully calibrated according to the sensitivity of the query and a global privacy budget $\epsilon$. Sensitivity here is defined as the maximum amount of change in the query answer considering all neighboring datasets, \emph{i.e.}, datasets differing in the data of one individual (one row), or equivalently having Hamming distance of one. One limitation of this definition is that it provides the same level of protection for all attributes of the dataset, i.e., all elements in the data universe $\mathcal{X}$.

In many scenarios, a more flexible notion of neighboring datasets may be more useful. For instance, in some domains it might be more natural to measure the distinguishability between two datasets by some generic metric $d_\mathcal{X} : \mathcal{X} \times \mathcal{X} \rightarrow \mathbb{R}_+$ instead of just Hamming distance. A case in point is  location-based systems where it might be acceptable to disclose coarse-grained information about an individual's location instead of his/her exact location. In this case, geographical distance would be an appropriate measure of distinguishability \cite{andres2013geo}. There are other scenarios where some attributes of the dataset may need more protection than others, and vice versa. As an example, consider a classification problem with instance space $\mathcal{X} \subset \mathbb{R}^d$ where specific features of $\mathcal{X}$ are more sensitive than others (maybe due to fairness requirements \cite{dwork2012fairness}). In this case, $d_{\mathcal{X}}\br{u,v} = \sum_{i=1}^{d}{\epsilon_i \ind{u_i \neq v_i}}, \forall{u,v \in \mathcal{X}}$ might be a reasonable choice for the metric, where $\epsilon_i$ is the privacy budget for the $i$th feature.\footnote{Note that $\ind{P} = 1$ if predicate $P$ is true.}

In the applications mentioned above, standard differential privacy (with a global privacy budget) is too strong privacy guarantee and as a result compromises much in utility. To address this limitation, several relaxations of differential privacy have been proposed recently \cite{chatzikokolakis2013broadening,he2014blowfish}. Despite the existence of these alternative proposals, they have gained little traction among practitioners. Part of the reason, we believe, is the dearth of standard procedures to develop algorithms satisfying these alternative definitions, as compared to differential privacy. In this work, we attempt to bridge this gap, by building on the privacy notion of \textit{$d_\mathcal{X}$-privacy}, introduced in \cite{chatzikokolakis2013broadening}. Intuitively, $d_\mathcal{X}$-privacy allows specifying a separate privacy budget for each pair of elements $u, v$ in the data universe $\mathcal{X}$, given by the value $d_\mathcal{X}(u, v )$. 

We propose a generic strategy to tailor any differentially private mechanism to satisfy $d_\mathcal{X}$-privacy for linear queries. Given \emph{any} data universe $\mathcal{X}$ and \emph{any} choice of the metric $d_\mathcal{X}$, the procedure shows how to convert a differentially private mechanism into its $d_\mathcal{X}$-private counterpart tailored to the given utility measure. The resulting mechanism then provides a better trade-off between utility and privacy. Our main contributions are summarized as follows:
\begin{itemize}
\item We describe a meta procedure (for any metric) to tailor any existing differentially private mechanism into a $d_{\mathcal{X}}$-private variant for the case of linear queries. The main challenge is that the privacy budget, \emph{i.e.}, $d_\mathcal{X}$, is specified on the input universe $\mathcal{X}$, whereas the noise is added to the query response which belongs to the outcome space. 
\item The main component of our approach is a pre-processing optimization step, which depends on the utility measure of interest, to choose model parameters of the mechanism. We provide explicit formulation of this pre-processing optimization problem for some commonly used utility measures (under any $d_\mathcal{X}$-metric). In general, these problems are non-convex and computationally challenging. But we show that for certain loss functions the optimization problem can be approximately solved using heuristic approaches  (cf. Algorithm~\ref{algo:psa} in Section~\ref{sec:syn-experiments-multi}).
\item Based on the meta procedure, we describe $d_{\mathcal{X}}$-private variants of several well-known online and offline $\epsilon$-differentially private algorithms. In particular, we illustrate $d_{\mathcal{X}}$-private variants of the Laplace~\cite{dwork2006calibrating} and Exponential~\cite{mcsherry2007mechanism} mechanisms, as well as the SmallDB~\cite{blum2013learning} and MWEM~\cite{hardt2012simple} mechanisms, the latter two being mechanisms for releasing synthetic data. We remark that the choice of these algorithms is merely for demonstration. Our meta procedure can similarly be applied to other differentially private mechanisms.
\item We demonstrate the effectiveness of $d_{\mathcal{X}}$-privacy in terms of utility, by evaluating the proposed $d_{\mathcal{X}}$-private Laplace mechanism on both synthetic and real datasets using a set of randomly generated linear queries. In both cases we define the $d_{\mathcal{X}}$ metric as the Euclidean distance between elements in the data universe. Our results show that the utility from the $d_{\mathcal{X}}$-private Laplace mechanism is higher than its \textit{vanilla} counterpart, with some specific queries showing significant improvement.
\item  Finally, we demonstrate how $d_{\mathcal{X}}$-privacy generalizes and relates to other alternative privacy notions proposed in literature by extending our techniques to  Blowfish \cite{he2014blowfish} privacy (without constraints).
\item Our work is the first to propose $d_{\mathcal{X}}$-private mechanisms for linear queries over histograms in the centralized model. This is in contrast to the most related work to ours, i.e.,~\cite{andres2013geo} where the authors focus on location-based systems in the local model, and~\cite{chatzikokolakis2013broadening} which only considers universally optimal mechanisms under some specific $d_{\mathcal{X}}$ metrics.  
\end{itemize}

\section{Background and $d_{\mathcal{X}}$-Privacy}
\label{sec:preliminaries}
This section gives the background on differential privacy and associated concepts of linear queries, sensitivity, and utility. We also introduce $d_{\mathcal{X}}$-privacy and its relation to other privacy notions. 
\subsection{Notation}
Let $\bs{n} := \bc{1,\dots , n}$ for $n \in \mathbb{N}$, and $\mathbb{R}_+ := [0,\infty)$. We write $\ind{P} = 1$ if $P$ is true and $\ind{P} = 0$ otherwise. Let $x_i$ denote the $i$th coordinate of the vector $x$, and $A_{i,:}$ denote the $i$th row of the matrix $A$. We denote the inner product of two vectors $x,y \in \mathbb{R}^n$ by $\ip{x}{y}$. The $k$-element vector of all ones is denoted $\mathbf{1}_k := \br{1,\dots , 1}^\top$. For two vectors $a,b \in \mathbb{R}^n$, the operation $a \odot b$ represents element-wise multiplication. For $a \in \mathbb{R}^n$ and $B \in \mathbb{R}^{n \times d}$, the operation $a \odot B$ represents row-wise scalar multiplication of $B$ by the associated entry of $a$. For a vector $a \succeq 0$ represents that the vector is element-wise non-negative. Hamming distance is defined as $\norm{x - y}_H := \sum_{i=1}^{n}{\ind{x_i \neq y_i}}$. The $\ell_p$-norms are denoted by $\norm{\cdot}_p$. For a matrix $A$, define $\norm{A}_{p} := \br{\sum_i {\norm{A_{i,:}}_p^p}}^{1/p}$. 

\subsection{Differential Privacy}
Let $\mathcal{X}$ denote the data universe and $N := \abs{\mathcal{X}}$ its size. A database $D$ of $n$ rows is modelled as a histogram $x \in \mathbb{N}^N$ (with $\norm{x}_1 = n$), where $x_i$ encodes the number of occurrences of the $i$th element of the universe $\mathcal{X}$.\footnote{Note that each element of the universe is composed of attribute values from all attributes (columns) in the tuple-wise representation of the database.} Two neighboring databases $D$ and $D'$ (from $\mathcal{X}^n$) that differ in a single row ($\norm{D-D'}_H = 1$) correspond to two histograms $x$ and $x'$ (from $\mathbb{N}^N$) satisfying $\norm{x - x'}_1 = 2$. A mechanism $\mathcal{M} : \mathbb{N}^N \times \mathcal{Q} \rightsquigarrow \mathcal{Y}$ (where $\mathcal{Y}$ is the outcome space, and $\mathcal{Q}$ is the query class) is a randomized algorithm which takes a dataset $x \in \mathbb{N}^N$ and a query $q : \mathbb{N}^N \rightarrow \mathcal{Y}$, and answers with some $a \in \mathcal{Y}$. 

\begin{definition}[Differential Privacy, \cite{dwork2006calibrating}]
	A mechanism $\mathcal{M} : \mathbb{N}^N \times \mathcal{Q} \rightsquigarrow \mathcal{Y}$ is called $\epsilon$-differentially private if for all $x,x' \in \mathbb{N}^N$ such that $\norm{x - x'}_1 \leq 2$, for every $q \in \mathcal{Q}$, and for every measurable $S \subseteq \mathcal{Y}$, we have
	\[
	\P{\mathcal{M}\br{x,q} \in S} \leq \exp\br{\epsilon} \P{\mathcal{M}\br{x',q} \in S} .
	\]
\end{definition}

Here $\epsilon > 0$ is a parameter that measures the strength of the privacy guarantee (smaller $\epsilon$ being a stronger guarantee).

\subsection{$d_{\mathcal{X}}$-Privacy}
\label{sec:dx-privacy-def}

We consider a more flexible privacy notion, which is a particular case of the definition from \cite{chatzikokolakis2013broadening}, for statistical databases. Given a metric $d_{\mathcal{X}}$ on the data universe, a mechanism satisfies $d_{\mathcal{X}}$-privacy if the densities of the output distributions on input datasets $x,x' \in \mathbb{N}^N$ with $\norm{x - x'}_1 \leq 2$ and differing on $i,j$-th entries are pointwise within an $\exp\br{d_{\mathcal{X}} \br{i , j}}$ multiplicative factor of each other. 

\begin{definition}[$d_{\mathcal{X}}$-Privacy]
	\label{dx-privacy-def-hist}
	Let $d_{\mathcal{X}}: \bs{N} \times \bs{N} \rightarrow \mathbb{R}_+$ be the privacy budget (such that $d_{\mathcal{X}}\br{i,j} \geq 0$, $d_{\mathcal{X}}\br{i,j} = d_{\mathcal{X}}\br{j,i}$, $d_{\mathcal{X}}\br{i,i} = 0$, and $d_{\mathcal{X}}\br{i,j} \leq d_{\mathcal{X}}\br{i,k} + d_{\mathcal{X}}\br{k,j}$, $\forall{i,j,k \in \bs{N}}$) of the data universe $\mathcal{X}$. A mechanism $\mathcal{M}: \mathbb{N}^N \times \mathcal{Q} \rightsquigarrow \mathcal{Y}$ is said to be $d_{\mathcal{X}}$-private iff $\forall{x,x' \in \mathbb{N}^N}$ s.t. $\norm{x-x'}_1 \leq 2$, $x_i \neq x'_i$, and $x_j \neq x'_j$ (for some $i,j \in \bs{N}$), $\forall S \subseteq \mathcal{Y}$ and $\forall{q \in \mathcal{Q}}$ we have
	\[
	\frac{\P{\mathcal{M}\br{x,q} \in S}}{\P{\mathcal{M}\br{x',q} \in S}} ~\leq~ \exp\br{d_{\mathcal{X}} \br{i , j}} .
	\]
	When $d_{\mathcal{X}} \br{i , j} = \epsilon, \forall{i,j \in \bs{N}}$, we recover the standard $\epsilon$-differential privacy. 
\end{definition}

Most of the desirable properties of differential privacy is carried over to $d_{\mathcal{X}}$-privacy as well, with suitable generalization~\cite{chatzikokolakis2013broadening}.
\begin{fact}[Properties of $d_{\mathcal{X}}$-Privacy]
	$d_{\mathcal{X}}$-privacy satisfies the following properties:
	\begin{enumerate}
		\item Resistant to post-processing: If $\mathcal{M}:\mathbb{N}^N \times \mathcal{Q} \rightsquigarrow \mathcal{Y}$ is $d_{\mathcal{X}}$-private, and $f: \mathcal{Y} \rightarrow \mathcal{Y}'$ is any arbitrary (randomized) function, then the composition $f \circ \mathcal{M} : \mathbb{N}^N \times \mathcal{Q} \rightsquigarrow \mathcal{Y}'$ is also
		$d_{\mathcal{X}}$-private.
		\item Composability: Let $\mathcal{M}_i: \mathcal{X}^n \times \mathcal{Q}_i \rightsquigarrow \mathcal{Y}_i$ be a $d_{\mathcal{X}}^i$-private algorithm for $i \in \bs{k}$. If $\mathcal{M}_{\bs{k}}: \mathcal{X}^n \times \Pi_{i=1}^k{\mathcal{Q}_i} \rightsquigarrow \Pi_{i=1}^k{\mathcal{Y}_i}$ is defined to be: 
		\[
		\mathcal{M}_{\bs{k}} \br{x, \Pi_{i=1}^k{q_i}} = 
		\br{\mathcal{M}_{1}\br{x,q_1},\dots , \mathcal{M}_{k}\br{x,q_k}},
		\]
		then $\mathcal{M}_{\bs{k}}$ is $\sum_{i=1}^{k}{d_{\mathcal{X}}^i}$-private.
		\item Group privacy: If $\mathcal{M}:\mathbb{N}^N \times \mathcal{Q} \rightsquigarrow \mathcal{Y}$ is a $d_{\mathcal{X}}$-private mechanism and $x,x' \in \mathbb{N}^N$ satisfy $\norm{x - x'}_1 \leq k$ (with $k \geq 2$), then $\forall S \subseteq \mathcal{Y}$ and $\forall{q \in \mathcal{Q}}$ we have
		\[
		\frac{\P{\mathcal{M}\br{x,q} \in S}}{\P{\mathcal{M}\br{x',q} \in S}} ~\leq~ \exp\br{k \cdot \max_{i , j \in V}{d_{\mathcal{X}} \br{i , j}}} ,
		\]
		where $V$ is the set of indices in which $x$ and $x'$ differ.
	\end{enumerate}
\end{fact}
$d_\mathcal{X}$-privacy can naturally express indistinguishability requirements that cannot be represented by the standard notion of Hamming distance (between neighbouring datasets). But the metric $d_\mathcal{X}$ in the above definition must be appropriately defined to achieve meaningful privacy goals. We present some examples in Section~\ref{sec:discuss}. In this work, we mainly focus on how to convert an existing differentially private algorithm into $d_\mathcal{X}$-private equivalent, given an already appropriately defined $d_\mathcal{X}$-metric.

\subsection{Linear Queries and Sensitivity}
Our focus is on the trade-off between privacy and accuracy when answering a large number of \textit{linear queries} over histograms. Linear queries include some natural classes of queries such as range queries \cite{li2011efficient,li2012adaptive} and contingency tables \cite{barak2007privacy,fienberg2010differential}, and serve as the basis of a wide range of data analysis and learning algorithms (\emph{e.g.}, Perceptron, K-means clustering, PCA \cite{blum2005practical}). 
Formally, given a query vector $q \in \mathbb{R}^N$, a linear query over the dataset $x \in \mathbb{N}^N$ is defined as $q\br{x} = \ip{q}{x}$. A set of $k$ linear queries can be represented by a \textit{query matrix} $Q \in \mathbb{R}^{k \times N}$ with the vector $Q x \in \mathbb{R}^k$ giving the correct answers to the queries. 

For $d_{\mathcal{X}}$-privacy, we generalize the notion of \textit{global sensitivity}, defined in \cite{dwork2006calibrating}, as follows: 
\begin{definition}
	For $i,j \in \bs{N}$ (with $i \neq j$), the generalized global sensitivity of a query $q \in \mathcal{Q}$ (w.r.t. $\norm{\cdot}$) is defined as follows 
	\[
	\Delta_{\norm{\cdot}}^q \br{i,j} ~:=~ \max_{\substack{x,x' \in \mathbb{N}^N : \norm{x-x'}_1 \leq 2 ,\\ x_i \neq x'_i , x_j \neq x'_j \text{ for } i,j \in \bs{N}}}{\norm{q\br{x} - q\br{x'}}} .
	\]
	Also define $\Delta_{\norm{\cdot}}^q := \max_{i,j \in \bs{N}}{\Delta_{\norm{\cdot}}^q \br{i,j}}$ (the usual global sensitivity). When $\norm{\cdot} = \norm{\cdot}_p$, we simply write $\Delta_p^q$.
\end{definition}
Consider a multi-linear query $Q: \mathbb{N}^N \rightarrow \mathcal{Y} \subseteq \mathbb{R}^k$ defined as $Q \br{x} = Q x$, where $Q \in \mathbb{R}^{k \times N}$. Then the generalized global sensitivity of $Q$ (for $i,j \in \bs{N}$) is given by $\Delta_{\norm{\cdot}}^Q \br{i,j} = \norm{Q_{:,i} - Q_{:,j}}$. When $k=1$, \emph{i.e.}, for a single linear query $q \br{x} = \ip{q}{x}$, we have $\Delta_{\norm{\cdot}}^q \br{i,j} = \abs{q_i - q_j}$. Thus, \emph{the generalized notion is defined separately for each pair $i, j$ of elements in $\mathcal{X}$}.  

\subsection{Laplace and Exponential Mechanism}
\begin{definition}[Laplace Mechanism, \cite{dwork2006calibrating}]
    \label{def:dp-lap}
	For a query function $q:\mathbb{N}^N \rightarrow \mathbb{R}^k$ with $\ell_1$-sensitivity $\Delta^q_1$, Laplace mechanism
	will output
	\begin{equation}
	\label{multi-lap-eps-diff-eq}
	\textnormal{\textsf{Z}} ~=~ \mathcal{M}_{\mathrm{Lap}, \frac{\Delta^q_1}{\epsilon} \cdot \mathbf{1}_k}\br{x , q} ~:=~ q\br{x} + \br{\textnormal{\textsf{Y}}_1,\dots ,\textnormal{\textsf{Y}}_k} ,
	\end{equation}
	where $\textnormal{\textsf{Y}}_i \overset{\mathrm{iid}}{\sim} \mathrm{Lap}\br{\frac{\Delta^q_1}{\epsilon}}$, and $\mathrm{Lap}\br{\lambda}$ is a distribution with probability density function $f\br{x} = \frac{1}{2 \lambda} e^{- \frac{\abs{x}}{\lambda}} , \quad \forall{x \in \mathbb{R}}$.
\end{definition}

The Laplace mechanism satisfies $\epsilon$-differential privacy, but it satisfies $d_{\mathcal{X}}$-privacy only with $\epsilon \leq \min_{i,j \in \bs{N}}{d_{\mathcal{X}} \br{i,j}}$. This would result in large noise addition, and eventually unnecessary compromise on overall utility. 

Given some arbitrary range $\mathcal{R}$, the exponential mechanism is defined with respect to some utility function $u: \mathbb{N}^N \times \mathcal{R} \rightarrow \mathbb{R}$, which maps database/output pairs to utility scores. The sensitivity notion that we are interested here is given by:
\begin{definition}
	For $i,j \in \bs{N}$ (with $i \neq j$) and $u: \mathbb{R}^N \times \mathcal{R} \rightarrow \mathbb{R}$, the generalized utility sensitivity is defined as follows
	\[
	\Delta u \br{i,j} := \max_{r \in \mathcal{R}}{\max_{\substack{x,x' \in \mathbb{N}^N : \norm{x-x'}_1 \leq 2 ,\\ x_i \neq x'_i , x_j \neq x'_j \text{ for } i,j \in \bs{N}}}{\abs{u\br{x,r} - u\br{x',r}}}} .
	\]
	Also define $\Delta u := \max_{i,j \in \bs{N}}{\Delta u \br{i,j}}$.
\end{definition}
Formally, the exponential mechanism is:
\begin{definition}[The Exponential Mechanism, \cite{mcsherry2007mechanism}]
	The exponential mechanism $\mathcal{M}_{\mathrm{Exp}, \frac{\Delta u}{\epsilon}}\br{x , u}$ selects and outputs an element $r \in \mathcal{R}$ with probability proportional to $\exp\br{\frac{\epsilon u\br{x,r}}{2 \Delta u}}$.
\end{definition}

The exponential mechanism satisfies $\epsilon$-differential privacy. The resulting mechanism also satisfies $d_{\mathcal{X}}$-privacy only if we set $\epsilon \leq \min_{i,j \in \bs{N}}{d_{\mathcal{X}} \br{i,j}}$. 

\subsection{Utility}
In the differential privacy literature, the performance of a mechanism is usually measured in terms of its worst-case total expected \textit{error}, defined as follows:
\begin{definition}[Error]
	Let $q: \mathbb{N}^N \rightarrow \mathcal{Y} \subseteq \mathbb{R}^k$ and $\ell: \mathbb{R}^k \times \mathbb{R}^k \rightarrow \mathbb{R}_+$. We define the $\ell$-error of a mechanism $\mathcal{M}: \mathbb{N}^N \times \mathcal{Q} \rightsquigarrow \mathcal{Y}$ as
	\begin{equation}
		\label{error-def-eq}
		\mathrm{err}_{\ell}\br{\mathcal{M},q} = \sup_{x \in \mathbb{N}^N}{\Ee{\textnormal{\textsf{Z}} \sim \mathcal{M}\br{x,q}}{\ell\br{\textnormal{\textsf{Z}},q\br{x}}}} .
	\end{equation}
	Here the expectation is taken over the internal coin tosses of the mechanism itself.
\end{definition}
In this paper, we are mainly interested in the worst case expected $\ell_p$-error defined by 
\[
\ell_p \br{y,\hat{y}} ~:=~ \norm{y - \hat{y}}_p ~=~ \br{\sum_{i=1}^{k}{\abs{y_i - \hat{y}_i}^p}}^{\frac{1}{p}} ,
\]
for $p \in \bc{1,2,\infty}$, and $\ell_2^2$-error (given by $\ell_2^2 \br{y,\hat{y}} := \norm{y - \hat{y}}_2^2$). It is also common to analyze high probability bounds on the accuracy of the privacy mechanisms.	
\begin{definition}[Accuracy]
	Given a mechanism $\mathcal{M}: \mathbb{N}^N \times \mathcal{Q} \rightsquigarrow \mathcal{Y}$, query $q: \mathbb{N}^N \rightarrow \mathcal{Y} \subseteq \mathbb{R}^k$, sensitive dataset (histogram) $x \in \mathbb{N}^N$, and parameters $\alpha > 0$ and $\beta \in \br{0,1}$, the mechanism $\mathcal{M}$ is $\br{\alpha ,\beta}$-accurate for $q$ on $x$ under the $\norm{\cdot}$-norm if $
	\P{\norm{\mathcal{M}\br{q,x} - q\br{x}} \geq \alpha} \leq \beta$
	where $\norm{\cdot}$-norm can be any vector norm definition. In our analysis, we consider the $\norm{\cdot}_1$-norm and the $\norm{\cdot}_{\infty}$-norm. 
\end{definition}

\section{$d_{\mathcal{X}}$-Private Mechanisms for Linear Queries}
\label{sec:mech-linear-query}
In this section, we design $d_{\mathcal{X}}$-private mechanisms by extending some of the well known $\epsilon$-differentially private (noise adding) mechanisms. Before delving into our approach, we exemplify potential technical issues when defining a variable privacy budget across attributes, and how $d_\mathcal{X}$-privacy provides a solution. 
\begin{example}
\label{ex:example}
Consider a simple domain $\mathcal{X}$ with three binary attributes described below.

\begin{center}
\begin{tabular}{ c|c|c } 
    Gender & Native & Age \\
 \hline\hline
 Male (\texttt{M}) & Yes (\texttt{Y}) & Above 18 (\texttt{A}) \\ 
 Female (\texttt{F}) & No (\texttt{N}) & Below 18 (\texttt{B}) \\ 
\end{tabular}
\end{center}

Figure~\ref{fig:ex-hist} shows an example dataset $x$ (as a histogram) from this domain, where we have used abbreviated attribute value names to describe each element $x_i$, $i \in [N] = 
[8]$. Assume that the attribute value ``Native = {\tt{Y}}'' is considered sensitive and all other values non-sensitive. Using differential privacy, the data custodian might wish to use some 
privacy budget $\epsilon = \epsilon_0$ for the attribute value {\tt{Y}}, and $\epsilon = \infty$ for all other attribute values. Any linear query $q \in \mathbb{R}^8$ with a non-zero value for 
coordinates 1, 2, 5 or 6 (containing attribute value {\tt{Y}}), would be answered with the budget $\epsilon_0$, and all remaining queries with budget $\infty$ (i.e., noiseless answers). While 
this may sound reasonable, notice that an analyst can obtain noiseless answer to the query ({\tt{MNA}}, {\tt{MNB}}, {\tt{FNA}}, {\tt{FNB}}) = ({\tt{N}}),\footnote{We are omitting zeroed coordinates in this equivalent 
notation.} and get the answer to ({\tt{MYA}}, {\tt{MYB}}, {\tt{FYA}}, {\tt{FYB}}) = ({\tt{Y}}) without noise (since $\norm{x}_1 = n$ is assumed to be publicly known). This simple example shows why attribute-wise privacy budget allocation should obey properties of a distance metric.

\pgfplotsset{width=9cm, height=3cm}

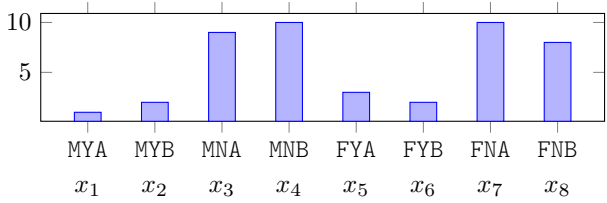
\begin{figure}
    \centering
    \begin{tikzpicture}
    \begin{axis}[
            ybar,
            xtick=data,
            xticklabels from table={mydata.dat}{labels},
            xticklabel style={align=center}
        ]
        \addplot table [x=interval,y=counts]{mydata.dat};
    \end{axis}
    \end{tikzpicture}
    \caption{Histogram representation of the dataset $x$ used in Example~\ref{ex:example}; horizontal axis contains domain elements, while the vertical axis shows counts.}
    \label{fig:ex-hist}
\end{figure}

We can solve this using $d_{\mathcal{X}}$-privacy as follows. Denote by $\epsilon(X)$ the privacy budget allocated to attribute value $X$. Then, set $\epsilon({\tt{Y}}) = \epsilon_0$, and $\epsilon(X) = \epsilon_1 > \epsilon_0$, for all $X \neq {\tt{Y}}$. Denote each of the $x_i$'s as $x_i = X_i^{(1)}X_i^{(2)}X_i^{(3)}$, where $X_i^{(k)}$ denotes the $k$th attribute value of $x_i$. Finally, we define 
\begin{equation}
\label{eq:ex-dx}
d_{\mathcal{X}}(i, j) = \sum_{k = 1}^3 \min\{\epsilon(X_i^{(k)}),\epsilon(X_j^{(k)})\} \ind{X_i^{(k)} \neq X_j^{(k)}}.
\end{equation}
For instance, $d_{\mathcal{X}}(1, 1) = 0$, $d_{\mathcal{X}}(1, 2) = \epsilon_1 \cdot 0 + \epsilon_0 \cdot 0 + \epsilon_1 \cdot 1 = \epsilon_1$,  $d_{\mathcal{X}}(1, 3) = \epsilon_0$, and 
$d_{\mathcal{X}}(1, 8) = \epsilon_0 + 2\epsilon_1$. It is easy to verify that this is indeed a distance metric. As we will show in Section~\ref{sec:syn-experiments-single}, given any linear query 
$q \in \mathbb{R}^8$, one way to answer through our framework is to let $c = \max_{i, j} \frac{|q_i - q_j |}{d_{\mathcal{X}}(i, j)}$, and then add Laplace noise of scale $c$ to the answer. Thus, 
for instance, if $q = ({\tt{MNA}}, {\tt{MNB}}, {\tt{FNA}}, {\tt{FNB}}) = ({\tt{N}})$, then the maximum is achieved at $i = 1, j = 3$ (not uniquely), which gives us $c = 1/\epsilon_0$. We have already seen that 
this is indeed a sensitive query, and is equivalent to answering the query $({\tt{Y}})$. Thus, the scale of noise is justified. Likewise, for the query $q = (1, 1, 1, 1, 0, 0, 0, 0) = ({\tt{M}})$, 
which is not considered sensitive, the maximum is achieved at $i = 1, j = 5$ (again, not uniquely), giving us $c = 1 / \epsilon_1$. Thus, less noise is added to the answer to this 
query. Finally, notice that with the convention $a + \infty = \infty$ for all $a \in \mathbb{R}^+$, if we set $\epsilon_1 = \infty$, Eq.~\ref{eq:ex-dx} still defines a distance metric. Thus we can 
even get noiseless answers to the ``non-sensitive'' queries.\qed
\end{example}

Consider a query $q : \mathbb{N}^N \rightarrow \mathbb{R}$, and the Laplace mechanism (Definition~\ref{def:dp-lap}). To achieve $\epsilon$-differential privacy, we compute the difference in the answers to the query over all neighbouring databases, given by the global sensitivity $\Delta_{\norm{\cdot}}^{q}$, and then add noise of scale $\Delta_{\norm{\cdot}}^{q}/\epsilon$. The corresponding quantity in $d_\mathcal{X}$-privacy is $\Delta_{\norm{\cdot}}^{q}(i, j)/d_\mathcal{X}(i, j)$; but this is potentially different for all $i, j \in [N]$. Therefore, to obtain a $d_\mathcal{X}$-private counterpart, we need to obtain a single optimum noise scale $c \ge 0$ subject to the constraints $c \ge \Delta_{\norm{\cdot}}^{q}(i, j)/d_\mathcal{X}(i, j)$, for all $i, j \in [N]$ (to guarantee privacy). Equivalently, if we introduce a parameter $q'$ and set it approximately equal to $q/c$, i.e., $q' \approx q/c$, then we perturb the answer to the \emph{scaled query} $c q' \approx q$ with Laplace noise of scale $c$  subject to the condition $\Delta_{\norm{\cdot}}^{q'}(i, j) \le d_\mathcal{X}(i, j)$, for all $i, j \in [N]$. The parameters $c$ and $q'$ can be optimized for a given measure of utility (within the privacy constraints imposed by the $d_\mathcal{X}$ metric). Note that the two parameters are not dependent on the input data, and can be optimized using pre-processing without compromising privacy. To summarize, given a query $q$, we have a two-step process: (a) obtain parameters $c$ and $q'$ through optimization (via pre-processing), (b) convert the $\epsilon$-differentially private Laplace mechanism into its $d_\mathcal{X}$-private counterpart by replacing input $(q, \Delta_{\norm{\cdot}}^{q}/\epsilon)$ by input $(cq', c)$. The exact form of the optimization problem depends on the utility measure under consideration. We generalize this procedure in the following.


Given a dataset $x \in \mathbb{N}^N$, and a query $q:\mathbb{N}^N \rightarrow \mathcal{Y} \subseteq \mathbb{R}^k$, our approach (meta procedure) to design a $d_{\mathcal{X}}$-private (noise adding) mechanism is as follows:

\begin{enumerate}
	\item Choose the (approximately optimal) model parameters $c \in \mathbb{R}^k$ and $q' : \mathbb{N}^N \rightarrow \mathbb{R}^k$ such that $\Delta_{\norm{\cdot}}^{q'} \br{i,j} \leq d_{\mathcal{X}}\br{i,j}$, $\forall{i,j \in \bs{N}}$, and $c \succeq 0$.
	\item Then use an existing $\epsilon$-differentially private mechanism with $\br{c \odot q', c}$ in place of $\br{q, \frac{\Delta_{\norm{\cdot}}^{q}}{\epsilon} \mathbf{1}_k}$.
\end{enumerate}

The model parameters $q'$ and $c$ are chosen by (approximately) solving the following pre-processing optimization problem (i.e. $\br{q',c} := F_{\text{pre-opt}}\br{q,n,d_{\mathcal{X}}\br{\cdot,\cdot},\ell}$):
\begin{equation}
\label{abstratc-pre-opt}
\begin{aligned}
& \underset{q',c}{\text{minimize}}
& & f_{\ell,\mathcal{M}}\br{q',c;q,n} \\
& \text{subject to}
& & \Delta_{\norm{\cdot}}^{q'} \br{i,j} \leq d_{\mathcal{X}}\br{i,j} , \quad \forall{i,j \in \bs{N}} \\
&&& c \succeq 0 ,
\end{aligned}
\end{equation}
where $f_{\ell,\mathcal{M}}\br{q',c;q,n}$ is a surrogate function of the utility measure that we are interested in. Note that this pre-processing optimization depends only on the data universe $\mathcal{X}$ (or $\bs{N}$), the query set $\mathcal{Q}$, and the database size $n$, but not on the dataset $x$. Thus we don't compromise any privacy during the optimization procedure. More over, we have to do the pre-processing optimization only once in an offline manner (for given $\mathcal{X}$, $\mathcal{Q}$, and $n$). The number of constraints in the optimization problem \eqref{abstratc-pre-opt} can be exponentially large ($\approx 2^N$), but depending on the structure of the $d_{\mathcal{X}}$-metric the constraint count can be significantly reduced (cf. Appendix~\ref{sec:stat-query}). 

\textbf{Remark 1:} Note that the pre-processing optimization problem \eqref{abstratc-pre-opt} to choose the model parameters of our new strategy depends on the 1) utility measure 2) $d_{\mathcal{X}}$-metric 3) $\epsilon$-differentially private mechanism that we want to transform. Every $\epsilon$-differentially private mechanism requires a separate (utility) analysis to derive $f_{\ell,\mathcal{M}}$ of \eqref{abstratc-pre-opt}. In this work, we consider fundamental online and offline (synthetic data generation) mechanisms (Laplace, Exponential, SmallDB, and MWEM) under specific utility measures and any metric. We can similarly extend our analysis to advanced $\epsilon$-differentially private mechanisms (for multi-linear queries) such as the Matrix~\cite{li2011efficient}, and $K$-norm~\cite{hardt2010geometry} mechanisms. We leave it as future work.

Next we apply the above described abstract meta procedure in extending some $\epsilon$-differential privacy mechanisms under different loss measures such as squared loss and absolute loss. We first show that the resulting mechanisms are in fact $d_{\mathcal{X}}$-private, and then we formulate the appropriate pre-processing optimization problems \eqref{abstratc-pre-opt} for them.

\subsection{$d_{\mathcal{X}}$-Private Laplace Mechanism}
For a given query $q : \mathbb{R}^N \rightarrow \mathcal{Y} \subset \mathbb{R}^k$ over the histogram $x \in \mathbb{R}^N$, consider the following variant of Laplace mechanism (with the model parameters $q' : \mathbb{R}^N \rightarrow \mathbb{R}^k$, and $c \in \mathbb{R}^k$ which depend on the utility function of the task):
\begin{equation}
\label{weight-laplace-multi}
\textsf{Z} = \mathcal{M}_{\mathrm{Lap},c}\br{x , c \odot q'} := c\odot q'\br{x} + \br{\textsf{Y}_1,\dots ,\textsf{Y}_k} ,
\end{equation}
where $\textsf{Y}_i \overset{\perp}{\sim} \mathrm{Lap}\br{c_i}$. When $q\br{x} = Q x$ (i.e., $q$ is a multi-linear query), we choose $Q' \in \mathbb{R}^{k \times N}$ and $c\in \mathbb{R}^k$ as the model parameters \emph{i.e.}, $q'\br{x} = Q' x$. Below we show that the above variant of Laplace mechanism satisfies $d_{\mathcal{X}}$-privacy under a sensitivity bound condition. 
\begin{theorem} 
	\label{laplace-privacy-prop}
	If $\Delta_1^{q'} \br{i,j} \leq d_{\mathcal{X}}\br{i,j}$, $\forall{i,j \in \bs{N}}$, then the mechanism $\mathcal{M}_{\mathrm{Lap},c}\br{\cdot , c \odot q'}$ given by \eqref{weight-laplace-multi} satisfies $d_{\mathcal{X}}$-privacy. 
\end{theorem} 

The sensitivity bound condition of the above theorem for a multi-linear query $Q'\br{x} = Q' x$ can be written as: $\Delta_1^{Q'} \br{i,j} = \norm{Q'_{:,i} - Q'_{:,j}}_1 \leq d_{\mathcal{X}}\br{i,j} , \quad \forall{i,j \in \bs{N}}$. The next theorem characterizes the performance of the $\mathcal{M}_{\mathrm{Lap},c}\br{\cdot , c \odot q'}$ mechanism under different choices of utility measures:
\begin{theorem}
	\label{laplace-utility-prop}
	Let $Q: \mathbb{R}^N \rightarrow \mathbb{R}^k$ be a multi-linear query of the form $Q\br{x} = Q x$, and let $\textnormal{\textsf{Z}} = \mathcal{M}_{\mathrm{Lap},c} \br{x,c\odot Q'} = c\odot Q'x + \textnormal{\textsf{Y}}$ with $\textnormal{\textsf{Y}}_i \overset{\perp}{\sim} \mathrm{Lap}\br{c_i}$. 
	\begin{enumerate}
		\item When $\ell_2^2 \br{y,y'} = \norm{y - y'}_2^2$, we have
		\begin{align*}
		    \mathrm{err}_{\ell_2^2}\br{\mathcal{M}_{\mathrm{Lap},c} \br{\cdot,c\odot Q'},Q} ~\leq~& 2 n^2 \norm{c\odot Q' - Q}_{2}^2 \\
		    & \quad + 4 \norm{c}_2^2 ,
		\end{align*}
		where $\mathrm{err}_{\ell}\br{\mathcal{M},Q}$ is defined in \eqref{error-def-eq}.
		\item When $\ell_p \br{y,y'} = \norm{y - y'}_p$, we have
		\begin{align*}
		    \mathrm{err}_{\ell_p}\br{\mathcal{M}_{\mathrm{Lap},c} \br{\cdot,c\odot Q'},Q} ~\leq~& n \norm{c\odot Q' - Q}_{p}  \\ 
		    & + \Ee{\textnormal{\textsf{Y}}_i \overset{\perp}{\sim} \mathrm{Lap}\br{c_i}}{\norm{\textnormal{\textsf{Y}}}_p} .
		\end{align*}
		Note that $\Ee{\textnormal{\textsf{Y}}_i \overset{\perp}{\sim} \mathrm{Lap}\br{c_i}}{\norm{\textnormal{\textsf{Y}}}_1} ~=~ \norm{c}_1$.
		\item $\forall{\delta \in (0,1]}$, with probability at least $1-\delta$ we have 
		\[
		\norm{Q x - \textnormal{\textsf{Z}}}_\infty  \leq n \norm{c\odot Q' - Q}_{\infty} + \ln\br{\frac{k}{\delta}} \cdot \norm{c}_{\infty} .
		\]
	\end{enumerate}
\end{theorem}

Proofs of Theorem~\ref{laplace-privacy-prop} and~\ref{laplace-utility-prop} are given in Appendix~\ref{app:lap-mech}. Based on the upper bounds that we obtained in the previous theorem, we can formulate the pre-processing optimization problem $F_{\text{pre-opt}}\br{q,n,d_{\mathcal{X}}\br{\cdot,\cdot},\ell}$ to select the model parameters $c$ and $Q'$ of the $\mathcal{M}_{\mathrm{Lap},c} \br{\cdot,c\odot Q'}$ mechanism as follows:
\begin{equation}
\label{opt-laplace-multi-linear-query}
\begin{aligned}
& \underset{Q',c}{\text{minimize}}
& & f_{\ell,\mathcal{M}_{\mathrm{Lap},c} \br{\cdot,c\odot Q'}}\br{Q',c;Q,n} \\
& \text{subject to}
& & \norm{Q'_{:,i} - Q'_{:,j}}_1 \leq d_{\mathcal{X}} \br{i,j} ,  \forall{i,j \in \bs{N}} \\
&&& c\succeq 0.
\end{aligned}
\end{equation}
The objective function of the above optimization problem depends on the utility function that we are interested in. For example, when $\ell_2^2 \br{y,y'} = \norm{y - y'}_2^2$, we can choose $f_{\ell_2^2,\mathcal{M}_{\mathrm{Lap},c} \br{\cdot,c\odot Q'}}\br{Q',c;Q,n} = n^2 \norm{c\odot Q' - Q}_{2}^2 + 2 \norm{c}_2^2$. In summary, the $d_{\mathcal{X}}$-private Laplace mechanism, under $\ell_2^2$-error function, can be described as follows:

\begin{enumerate}
	\item Choose the model parameters $\br{Q',c}$ by approximately solving the pre-processing optimization problem $F_{\text{pre-opt}}\\ \br{Q,n,d_{\mathcal{X}}\br{\cdot,\cdot},\ell_2^2}$ given by \eqref{opt-laplace-multi-linear-query}. 
	\item Release the response $\textsf{Z} = \mathcal{M}_{\mathrm{Lap},c} \br{x,c\odot Q'}$ given by \eqref{weight-laplace-multi}.
\end{enumerate}

Observe that, when $d_{\mathcal{X}}\br{i,j} = \epsilon, \forall{i,j \in \bs{N}}$, the choices $c_i = c= \frac{\Delta^Q_1}{\epsilon}, \forall{i \in \bs{k}}$ and $Q' = \frac{1}{c} Q$ satisfy the constraints of the optimization problem~\eqref{opt-laplace-multi-linear-query} under squared loss. In fact these choices correspond to the standard Laplace mechanism $\mathcal{M}_{\mathrm{Lap}, \frac{\Delta^q_1}{\epsilon} \cdot \mathbf{1}_k}$, and thus our framework is able to recover standard $\epsilon$-differential privacy mechanisms as well.

The optimization problem~\eqref{opt-laplace-multi-linear-query} is in general non-convex, which is indeed hard to optimize. However, certain instances of this problem (instantiated by the utility function) allow efficient solutions in light of recent results. We discuss this in Appendix~\ref{sec:biopt}. Also note that even if globally optimal solutions are infeasible to obtain, an approximate solution might still yield good utility in practice. We show this in Section~\ref{sec:experiments}. 


\subsection{$d_{\mathcal{X}}$-Private Exponential Mechanism}
For a given utility function $u : \mathbb{N}^N \times \mathcal{R} \rightarrow \mathbb{R}$ over the histogram $x \in \mathbb{R}^N$, consider the following variant of exponential mechanism (with the model parameters $u' : \mathbb{N}^N \times \mathcal{R} \rightarrow \mathbb{R}$, and $c \in \mathbb{R}$ which will be chosen later based on the utility function): 
\begin{definition}
	The mechanism $\mathcal{M}_{\mathrm{Exp},c}\br{x,u'}$ selects and outputs an element $r \in \mathcal{R}$ with probability proportional to $\exp\br{\frac{u'\br{x,r}}{2 c}}$.
\end{definition}
Here we note that for ease of presentation, we do not consider using $c_r \in \mathbb{R}$ for each $r \in \mathcal{R}$. The following theorem provides a sufficient condition for the above mechanism to satisfy $d_{\mathcal{X}}$-privacy.
\begin{theorem} 
	\label{exp-privacy-theorem}
	If $\Delta u' \br{i,j} \leq cd_{\mathcal{X}}\br{i,j}, \forall{i,j \in \bs{N}}$, then the mechanism $\mathcal{M}_{\mathrm{Exp},c}\br{\cdot,u'}$ satisfies $d_{\mathcal{X}}$-privacy. 
\end{theorem} 

For a given histogram $x$ and a given utility measure $u: \mathbb{R}^N \times \mathcal{R} \rightarrow \mathbb{R}$, let $\star_u \br{x} = \max_{r \in \mathcal{R}}{u\br{x,r}}$ denote the maximum utility score of any element $r \in \mathcal{R}$ with respect to histogram $x$. Below we generalize the Theorem 3.11 from \cite{dwork2014algorithmic}:
\begin{theorem}
	\label{exp-utility-theorem}
	Fixing a database $x$, let $\mathcal{R}_{\star_{u'}} = \bc{r \in \mathcal{R} : u'\br{x,r} = \star_{u'} \br{x}}$
	denote the set of elements in $\mathcal{R}$ which attain utility score $\star_{u'} \br{x}$. Also define $\delta_{u,u'} := \max_{x,r}{\abs{u\br{x,r} - u'\br{x,r}}}$. Then for $\textnormal{\textsf{Z}} = \mathcal{M}_{\mathrm{Exp},c}\br{x , u'}$, with probability at least $1 - e^{-t}$, we have
	\begin{equation*}
	u \br{x, \textnormal{\textsf{Z}}} > \delta_{u,u'} + \star_{u'} \br{x} - 2 c\bc{\ln\br{\frac{\abs{\mathcal{R}}}{\abs{\mathcal{R}_{\star_{u'}}}}} + t} .
	\end{equation*}
	Since we always have $\abs{\mathcal{R}_{\star_{u'}}} \geq 1$, we get
	\[
	\P{u \br{x, \textnormal{\textsf{Z}}} \leq \delta_{u,u'} + \star_{u'} \br{x} - 2 c\bc{\ln\br{\abs{\mathcal{R}}} + t}} \leq e^{-t} . 
	\]
\end{theorem}

The proofs of the above two theorems are given in Appendix~\ref{app:exp-mech}. The exponential mechanism is a natural building block for designing complex $\epsilon$-differentially private mechanisms. Next we consider two data release mechanisms (i.e., offline synthetic data generation mechanisms) which use the Laplace and/or the expoential mechanism as building blocks. These are the $d_\mathcal{X}$-private variants of the small database mechanism \cite{blum2013learning}, and multiplicative weights exponential mechanism \cite{hardt2012simple}.


\subsection{$d_{\mathcal{X}}$-Private Small Database Mechanism}

Here we consider the problem of answering a large number of real valued linear queries $q:\mathbb{N}^N \rightarrow \mathbb{R}$ of the form $q\br{x} = \ip{q}{x}$ (where $q \in \mathbb{R}^N$, and $x \in \mathbb{N}^N$) from class $\mathcal{Q}$ via synthetic histogram/database release. For this problem \cite{blum2013learning} have proposed and studied a simple $\epsilon$-differentially private small database mechanism, which is an instantiation of exponential mechanism. They have used a utility function $u: \mathbb{N}^N \times \mathcal{R} \rightarrow \mathbb{R}$ (with $\mathcal{R} = \bc{y \in \mathbb{N}^N : \norm{y}_1 = \frac{\log{\abs{\mathcal{Q}}}}{\alpha^2}}$) defined as $u\br{x,y} := - \max_{q \in \mathcal{Q}}{\abs{q\br{x} - q\br{y}}}$.

Now we extend the mechanism developed in \cite{blum2013learning} to obtain a $d_{\mathcal{X}}$-private version of it using the model parameters $\mathcal{Q}'$ and $c \in \mathbb{R}$ (which are determined later). Algorithm~\ref{algo:small-db} is a modified version of Algorithm 4 from \cite{dwork2014algorithmic}, where the transformation from $\mathcal{Q}$ to $\mathcal{Q}'$ is one-to-one (thus we have $\abs{\mathcal{Q}'} = \abs{\mathcal{Q}}$). When answering a query $q \in \mathcal{Q}$ over $x$, we need to output $cq'\br{y}$ where $q' \in \mathcal{Q}'$ is the matching element of $q$ and $y$ is the output of the $d_{\mathcal{X}}$-private small database mechanism (Algorithm~\ref{algo:small-db}). The following theorem provides the $d_{\mathcal{X}}$-privacy characterization of the small database mechanism.

\begin{algorithm}[h]
	\caption{{\tt{Small Database Mechanism}}~\cite{blum2013learning}: SmallDB($x,\mathcal{Q}',c,\alpha$)}
	\label{algo:small-db}
	\begin{algorithmic}
		\STATE {\bfseries Let} $\mathcal{R} \leftarrow \bc{y \in \mathbb{N}^N : \norm{y}_1 = \frac{\log{\abs{\mathcal{Q}'}}}{\alpha^2}}$
		\STATE {\bfseries Let} $u': \mathbb{N}^N \times \mathcal{R} \rightarrow \mathbb{R}$ be defined to be: 
		\begin{equation}
			\label{small-db-utility-func}
			u'\br{x,y} ~:=~ - c\max_{q' \in \mathcal{Q}'}{\abs{q'\br{x} - q'\br{y}}} . 
		\end{equation}
		\STATE {\bfseries Sample And Output} $y \in \mathcal{R}$ with the mechanism $\mathcal{M}_{\mathrm{Exp},c}\br{x,u'}$
	\end{algorithmic}
\end{algorithm}

\begin{theorem}
	\label{small-db-privacy-prop}
	If $\abs{q'_i - q'_j} \leq d_{\mathcal{X}}\br{i,j}, \forall{i,j \in \bs{N}} \text{ and } \\ \forall{q' \in \mathcal{Q}'}$, then the small database mechanism is $d_{\mathcal{X}}$-private. 
\end{theorem}

The following proposition and theorem characterize the performance of the $d_{\mathcal{X}}$-private small database mechanism.
\begin{proposition}[Proposition 4.4, \cite{dwork2014algorithmic}]
	\label{utility-smalldb-prop}
	Let $\mathcal{Q}$ be any class of linear queries. Let $y$ be the
	database output by $\mathrm{SmallDB} \br{x,\mathcal{Q}',c,\alpha}$. Then with probability $1-\beta$:
	\begin{align*}
	    \max_{q \in \mathcal{Q}}{\abs{q\br{x} - cq'\br{y}}} ~\leq~& n \max_{q \in \mathcal{Q}}\norm{q-cq'}_\infty + \alpha n   \\
	    & + 2 c\bc{\frac{\log N \log \abs{\mathcal{Q}}}{\alpha^2} + \log\br{\frac{1}{\beta}}} .
	\end{align*}
\end{proposition}

\begin{theorem}[Theorem 4.5, \cite{dwork2014algorithmic}]
	\label{small-db-utitlity-theorem}
	By the appropriate choice of $\alpha$, letting $y$ be the database
	output by $\mathrm{SmallDB}\br{x,\mathcal{Q}',c,\frac{\alpha}{2}}$, we can ensure that with probability $1-\beta$:
	\begin{align*}
	    \max_{q \in \mathcal{Q}}{\abs{q\br{x} - cq'\br{y}}} 
	    ~\leq~& n \max_{q \in \mathcal{Q}}\norm{q-cq'}_\infty + \br{cn^2 \gamma}^{1/3} , 
	\end{align*}
	where $\gamma = 16 \log N \log \abs{\mathcal{Q}} + 4 \log\br{\frac{1}{\beta}}$. Equivalently, for any $c$ such that $c \leq \frac{\alpha^3 n}{\gamma}$ with probability $1-\beta$: $\max_{q \in \mathcal{Q}}{\abs{q\br{x} - cq'\br{y}}} \leq n \max_{q \in \mathcal{Q}}\norm{q-cq'}_\infty  + \alpha n$.
\end{theorem}

Proofs of these claims are given in Appendix~\ref{app:smalldb}. From the upper bound of the above theorem, the model parameters $\mathcal{Q}'$ and $c$ of the small database mechanism can be chosen through the following pre-processing optimization problem:
\begin{equation}
	\label{opt-smalldb}
	\begin{aligned}
		& \underset{{\mathcal{Q}}', c}{\text{minimize}}
		& & f \br{\mathcal{Q}',c;\mathcal{Q},n} \\
		& \text{subject to}
		& & \abs{q'_i - q'_j} ~\leq~ d_{\mathcal{X}} \br{i , j} , \, \forall{i,j \in \bs{N}, q' \in \mathcal{Q}'} \\
		&&& c\geq 0,
	\end{aligned}
\end{equation}
where $f \br{\mathcal{Q}',c;\mathcal{Q},n} = n \max_{q \in \mathcal{Q}}\norm{q-cq'}_\infty  + \br{cn^2 \gamma}^{1/3}$. Once again the optimization problem~\eqref{opt-smalldb} is non-convex. See Appendix~\ref{sec:biopt}, for a brief discussion on the (non-convex) pre-processing optimization problems \eqref{opt-laplace-multi-linear-query}, and \eqref{opt-smalldb}. 

\subsection{$d_{\mathcal{X}}$-Private Multiplicative Weights Exponential Mechanism}
\label{sec:mwem}
\begin{algorithm}[tb]
	\caption{\texttt{Multiplicative Weights Exponential Mechanism}~\cite{hardt2012simple}: MWEM($x, \mathcal{Q}', c, T$)}
	\label{algo:mwem}
	\begin{algorithmic}
		\STATE {\bfseries Input:} histogram $x$ over a universe $\bs{N}$, set $\mathcal{Q}'$ of linear queries, privacy
		parameter $c> 0$, and number of iterations $T \in \mathbb{N}$.
		\STATE {\bfseries Let} $n$ denote $\norm{x}_1$, the number of records in $x$. Let $y^0$ denote $n$ times the uniform distribution over $\bs{N}$.
		\FOR{$t=1,...T$}
		\STATE \begin{enumerate}
			\item \textit{Exponential Mechanism}: Sample a query $q'_t \in \mathcal{Q}'$ using the $\mathcal{M}_{\mathrm{Exp},2 c T}\br{x,u'_t}$ mechanism and the score function $u'_t : \mathbb{N}^N \times \mathcal{Q}' \rightarrow \mathbb{R}$ given by
			\begin{align*}
			u'_t \br{x,q'} ~:=~& c\abs{q'\br{y^{t-1}} - q'\br{x}} .
			\end{align*}
			\item \textit{Laplace Mechanism}: Let measurement $m_t = cq'_t\br{x} + \textnormal{\textsf{Y}}$ with $\textnormal{\textsf{Y}} \sim \mathrm{Lap}\br{2 c T}$.
			\item \textit{Multiplicative Weights}: Let $y^t$ be $n$ times the distribution whose entries satisfy $\forall i \in \bs{N}$,
			\[
			y^t_i \propto y^{t-1}_i \times \exp\br{\br{q'_t}_i \times \br{m_t - q'_t\br{y^{t-1}}}/2 n} .
			\]
		\end{enumerate}
		\ENDFOR
		\STATE {\bfseries Output} $y = \mathrm{avg}_{t < T} y^t$
	\end{algorithmic}
\end{algorithm}

As in the case of small database mechanism, here also we consider the problem of answering a large number of real valued linear queries in $d_{\mathcal{X}}$-private manner via synthetic histogram/database release. Algorithm~\ref{algo:mwem} is a simple modification of Algorithm 1 from \cite{hardt2012simple}. The following theorem provides the $d_{\mathcal{X}}$-privacy characterization of the MWEM mechanism.
\begin{theorem}
	\label{mwem-privacy-theorem}
	If $\abs{q'_i - q'_j} \leq d_{\mathcal{X}}\br{i,j}, \forall{i,j \in \bs{N}} \text{ and } \\ \forall{q' \in \mathcal{Q}'}$, then the MWEM mechanism is $d_{\mathcal{X}}$-private. 
\end{theorem}
The following theorem characterizes the performance of the MWEM mechanism.
\begin{theorem}[Theorem 2.2, \cite{hardt2012simple}]
	\label{mwem-utility-theorem}
	For any dataset $x$, set of linear queries $\mathcal{Q}$, $T \in \mathbb{N}$, and $c> 0$, with probability at least $1 - 2 T / \abs{\mathcal{Q}}$, MWEM produces $y$ such that 
	\begin{align*}
	& \max_{q \in \mathcal{Q}}\abs{cq'\br{y} - q\br{x}} \\
	~\leq~& 2 n \sqrt{\frac{\log N}{T}} + 10 T c\log\abs{\mathcal{Q}} + n \max_{q \in \mathcal{Q}} \norm{cq'-q}_\infty .
	\end{align*}
	By setting $2 n \sqrt{\frac{\log N}{T}} = 10 T c\log\abs{\mathcal{Q}}$, we get
	\begin{align*}
	& \max_{q \in \mathcal{Q}}\abs{cq'\br{y} - q\br{x}} \\
	~\leq~& n \max_{q \in \mathcal{Q}} \norm{cq'-q}_\infty + \frac{20}{5^{2/3}} \br{n^2 \log N \log\abs{\mathcal{Q}}}^{1/3} c^{1/3} .
	\end{align*}
\end{theorem}
Proofs of both theorems are given in Appendix~\ref{app:mwem}.
The model parameters $\mathcal{Q}'$ and $c$ of the MWEM mechanism can be chosen through the optimization problem~\ref{opt-smalldb} with 
\begin{align*}
f \br{\mathcal{Q}',c;\mathcal{Q},n} ~=~& n \max_{q \in \mathcal{Q}} \norm{cq'-q}_\infty \\
& \quad + \frac{20}{5^{2/3}} \br{n^2 \log N \log\abs{\mathcal{Q}}}^{1/3} c^{1/3}.
\end{align*}

\section{Experiments}
\label{sec:experiments}
In this section, we experimentally evaluate the effectiveness of our framework on both synthetic and real data. We will show that in many situations, we can drastically improve the accuracy of the noisy answers compared to the traditional differentially private mechanisms. The datasets considered in these experiments are geographic in nature. More specifically, for the ensuing experiments, the data universes considered consist of points in Euclidean space which allow an intuitive Euclidean distance-based $d_\mathcal{X}$-metric. Under this metric, fine-grained location information is protected while larger regions provide better utility. As stated in \cite{chatzikokolakis2013broadening}, when dealing with geographic locations, it might be acceptable to disclose the region of an individual. On the other hand, disclosing the precise location (town) of the individual is less desirable. Thus it is useful to have a distinguishability level that depends on the geographic distance. 

\subsection{Single Linear Queries over Synthetic Data}
\label{sec:syn-experiments-single}

\begin{figure*}[t]
	\begin{center}
	\captionsetup[subfigure]{font=scriptsize,labelfont=scriptsize}
	\begin{subfigure}[t]{0.48\textwidth}
		\centering
		\includegraphics[width=0.9\textwidth]{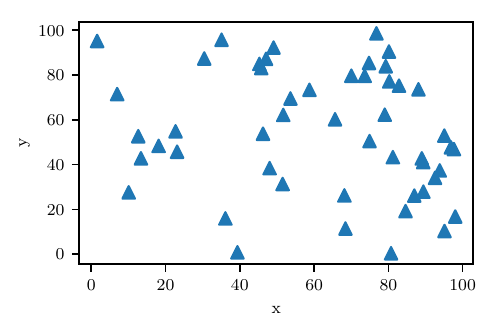}
		\caption{Elements of the data universe used in the synthetic data experiments}
		\label{fig:syn-single-data}
	\end{subfigure}
	\begin{subfigure}[t]{0.48\textwidth}
		\centering
		\includegraphics[width=0.9\textwidth]{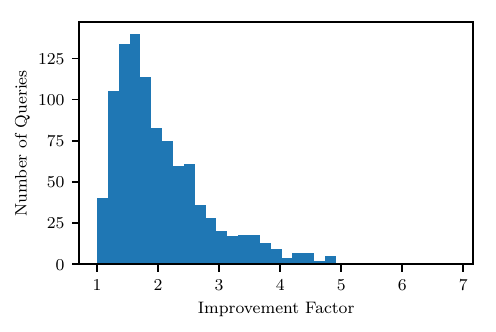}
		\caption{Histogram of the improvement factor (for 1000 random single linear queries)}
		\label{fig:syn-single-improvement}
	\end{subfigure}
	\begin{subfigure}[t]{0.48\textwidth}
		\centering
		\includegraphics[width=0.9\textwidth]{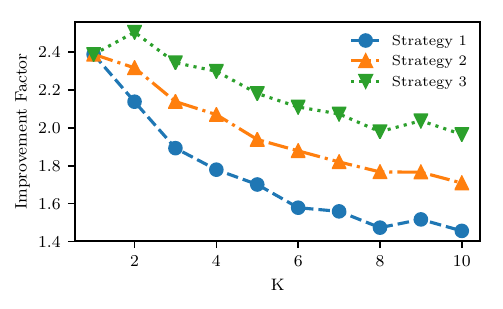}
		\caption{Improvement factor of multi-linear queries (random coefficients from real interval $[0, 1]$)}
		\label{fig:syn-multi-improvement-uniform}
	\end{subfigure}
	\begin{subfigure}[t]{0.48\textwidth}
		\centering
		\includegraphics[width=0.9\textwidth]{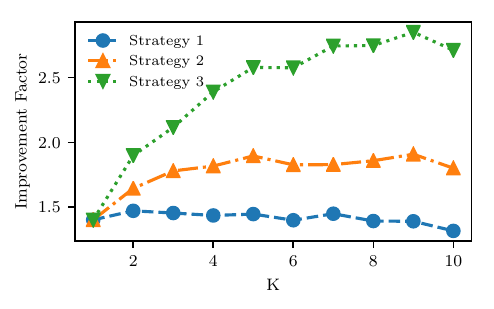}
		\caption{Improvement factor of multi-linear queries (random binary coefficients from $\bc{0, 1}$)}
		\label{fig:syn-multi-improvement-01}
	\end{subfigure}
	\caption{Synthetic experiment with $N=50$, and $d_{\mathcal{X}}$ metric defined based on Euclidean distance.}
	\label{fig:syn-single}
\end{center}
\end{figure*}
We first consider randomly generated single linear queries ($q:\mathbb{N}^N \rightarrow \mathbb{R}$), and compare the following two mechanisms: (a) the $\epsilon$-differentially private Laplace mechanism (with $\epsilon = \min_{i,j}{d_{\mathcal{X}}\br{i,j}}$): $\mathcal{M}_{\mathrm{Lap}, \frac{\Delta^q_1}{\epsilon}}\br{x , q} = q\br{x} + \textnormal{\textsf{Y}}$, where $\textnormal{\textsf{Y}} \sim \mathrm{Lap}\br{\frac{\Delta^q_1}{\epsilon}}$, and (b)
the $d_{\mathcal{X}}$-private Laplace mechanism (with the model parameters $c \in \mathbb{R}$ and $q' \in \mathbb{R}^N$): 
\[
\mathcal{M}_{\mathrm{Lap},c}\br{x , c q'} = c q'\br{x} + \textnormal{\textsf{Y}},
\]
where $\textnormal{\textsf{Y}} \sim \mathrm{Lap}\br{c}$, under the experimental setup given below.

\descr{Data and Privacy Metric:} 
We generate a random dataset (histogram) with $n = 10,000$ records from a data universe of size $N = 50$. We then randomly sample $N$ distinct two-dimensional points $\bc{\br{u_i,v_i}}_{i=1}^N$ from the set $S = \bs{0,100} \times \bs{0,100} \subseteq \mathbb{R}^2$, and associate each point $\br{u_i,v_i}$ with an element ($i \in \bs{N}$) of the data universe. Note that this simulates geographic locations over a region, e.g., user locations in a city.\footnote{Note that the data universe is fixed at $N = 50$ locations, each location exhibiting zero or more of the $n = 10,000$ records. Since this is a synthetic dataset, we choose a random data universe as well, by randomly sampling $N$ locations. In practice, these $N$ locations could be $N$ hotspots in a city. Privacy is provided for the $n=10,000$ subjects who can be in any of the $N$ locations in the data universe, with higher privacy for nearby locations.} The sampled data universe elements are shown in Figure~\ref{fig:syn-single-data}. We define the privacy metric $d_{\mathcal{X}}:\bs{N} \times \bs{N} \rightarrow \mathbb{R}$ based on the Euclidean distance (metric) on $2$-dimensional space. Specifically, for any $i,j \in \bs{N}$, define $d_{\mathcal{X}}\br{i,j} := \sqrt{\br{u_i - u_j}^2 + \br{v_i - v_j}^2}$.

\descr{Random Queries:} 
We evaluate the two mechanisms over $1000$ random single linear queries, where the query coefficients are randomly drawn from a uniform distribution over the real interval $\bs{0,1}$.

\descr{Performance Measure:}
We measure the individual performance of the mechanisms by the \textit{root mean squared error} (RMSE; between the private response and the actual output) on the above generated data, \emph{i.e.}, we consider the squared loss function $\ell \br{y,y'} = \norm{y - y'}_2^2$. Then the model parameters $c$ and $q'$ of the $d_{\mathcal{X}}$-private Laplace mechanism can be obtained by solving the following pre-processing optimization problem (for each query $q$):
\begin{equation*}
\begin{aligned}
& \underset{c,q'}{\text{minimize}}
& & f\br{c,q'} := n^2 \norm{cq' - q}_2^2 + 2 c^2 \\
& \text{subject to}
& & \abs{q'_{i} - q'_{j}} \leq d_{\mathcal{X}} \br{i,j} , \quad \forall{i,j \in \bs{N}} \\
&&& c > 0.
\end{aligned}
\end{equation*}

Since $n$ is very large, by fixing $c q' = q$ in the above problem, we obtain an approximately optimal (closed form) solution given by $c = \max_{i,j}\frac{\abs{q_{i} - q_{j}}}{d_{\mathcal{X}}\br{i,j}}$, and $q' = \frac{1}{c} q$.

\descr{Improvement Factor:}
We define another measure for cross-comparison of the two mechanisms. For a given single linear query $q$, the \emph{improvement factor} of the $d_{\mathcal{X}}$-private Laplace mechanism compared to the baseline ($\epsilon$-differentially private Laplace) mechanism is defined as $\mathrm{IF}\br{q} := \frac{\Delta^q_1 / \epsilon}{c}$. This factor is simply the ratio between the scales ($\lambda$) of the noise ($\mathrm{Lap}\br{\lambda}$) added by these two mechanisms. Then for each random query ($1000$ in total), we compute the improvement factor. The resulting values are presented in a histogram form in Figure~\ref{fig:syn-single-improvement}, where the $d_{\mathcal{X}}$-private mechanism exhibits significant improvement in utility compared to the baseline mechanism. Notice that IF does not depend on $\epsilon$, since $\epsilon$ is set to $\min_{i, j} d_{\mathcal{X}}(i, j)$, and $c$ is set to be inversely related to $\min_{i, j} d_{\mathcal{X}}(i, j)$. Thus, $\epsilon$ effecitvely ``cancels out'' in the definition of IF. As a result, the IF results in Figure~\ref{fig:syn-single-improvement} hold for any $\epsilon > 0$. We checked this for multiple values of $\epsilon$ and obtained similar plots. 

We note that IF is not a reasonable performance measure when the spread of the elements of the data universe is \emph{profoundly} non-uniform (\emph{e.g.}, two points are infinitesimally close to each other), in which case the traditional Laplace mechanism may get heavily penalized. But in both our real and synthetic data, the elements are (roughly) uniformly spread.

\subsection{Multi-Linear Queries over Synthetic Data}
\label{sec:syn-experiments-multi}

\begin{algorithm}[tb]
	\caption{\texttt{Parameter Selection Algorithm}: PSA($d_{\mathcal{X}}, Q$)}
	\label{algo:psa}
	\begin{algorithmic}
		\STATE {\bfseries Input:} privacy metric $d_{\mathcal{X}}$, and query matrix $Q \in \mathbb{R}^{K \times N}$.
		\STATE {\bfseries Let} $R = \mathbf{0}_K$, $T = \mathbf{1}_K$.
		\WHILE{$T \neq \mathbf{0}_K$}
		\STATE \begin{enumerate}
			\item $c'_k = \max_{i,j} \frac{\abs{Q_{k,i} - Q_{k,j}}}{d_{\mathcal{X}}\br{i,j}}$, $\forall{k \in \bs{K}}$. \COMMENT{near optimal scale of the noise for query $Q_{k,:}$, if the whole privacy budget is consumed by it.}
			\item $d_{\mathcal{X}}^k\br{i,j} = d_{\mathcal{X}}\br{i,j} \cdot \frac{\frac{1}{c'_k}\abs{Q_{k,i}-Q_{k,j}}}{\sum_{l=1}^{K}{\frac{1}{c'_l}\abs{Q_{l,i}-Q_{l,j}}}}$, $\forall{k \in \bs{K}}$, $\forall{i,j \in \bs{N}}$.\COMMENT{distribute the privacy budget between each query, based on $c'_k$'s.}
			\item $c_k = \max_{i,j} \frac{\abs{Q_{k,i} - Q_{k,j}}}{d_{\mathcal{X}}^k\br{i,j}}$, $\forall{k \in \bs{K}}$. \COMMENT{calculate the scale of the noise for each single linear query by considering the privacy budget allocated to them.}
			\item $d_{\mathcal{X}}\br{i,j} = d_{\mathcal{X}}\br{i,j} - \sum_{k=1}^{K}{\frac{1}{c_k}\abs{Q_{k,i}-Q_{k,j}}}$, $\forall{i,j \in \bs{N}}$. \COMMENT{calculate the remaining (total) privacy budget.}
			\item $T = \bs{\frac{1}{c_1},\dots,\frac{1}{c_K}}$, and $R = R + T$. \COMMENT{accumulate the share gained at this step.}
		\end{enumerate}
		\ENDWHILE
		\STATE {\bfseries Output:} $c = \bs{\frac{1}{R_1},\dots,\frac{1}{R_K}}$
	\end{algorithmic}
\end{algorithm}

Next we consider random multi-linear queries given by $Q \in \mathbb{R}^{K \times N}$, where we vary $K$ from $1$ to $10$. We consider the same data, privacy metric, and performance measure (squared loss) used in Section~\ref{sec:syn-experiments-single}. We consider two types of query matrices: the first type consists of matrices whose entries are drawn from a uniform distribution over the real interval $\bs{0, 1}$, and the second type has matrices whose entries are random binary numbers, i.e., elements of the set $\{0, 1\}$.

Again, we compare the $d_{\mathcal{X}}$-private Laplace mechanism \eqref{weight-laplace-multi} with the $\epsilon$-differentially private Laplace mechanism \eqref{multi-lap-eps-diff-eq}, with $\epsilon = \min_{i,j}{d_{\mathcal{X}}\br{i,j}}$. The model parameters $Q' \in \mathbb{R}^{K \times N}$ and $c\in \mathbb{R}^K$ of the $d_{\mathcal{X}}$-private Laplace mechanism \eqref{weight-laplace-multi} can be obtained from the optimization problem~\eqref{opt-laplace-multi-linear-query} with loss function $\ell \br{y,y'} = \norm{y - y'}_2^2$. Since $n$ is considerably large, by imposing the constraint $c \odot Q' = Q$, the resulting optimization problem can be written as follows (for each query $Q$):
\begin{equation*}
\begin{aligned}
& \underset{c}{\text{minimize}}
& & \norm{c}_2^2 = \sum_{k=1}^{K}{c_k^2} \\
& \text{subject to}
& & \sum_{k=1}^{K}{\frac{1}{c_k} \abs{Q_{k,i} - Q_{k,j}}} \leq d_{\mathcal{X}} \br{i,j} , \, \forall{i,j \in \bs{N}} \\
&&& c_k\geq 0, \, \forall{k \in \bs{K}}.
\end{aligned}
\end{equation*}
In particular, we consider the following three different strategies to choose $c\in \mathbb{R}^K$ (with $Q'_{k,:} = \frac{1}{c_k} Q_{k,:}, \forall{k \in [K]}$), which satisfy the constraints of the optimization problem above:
\begin{enumerate}
	\itemsep0em
	\item \emph{Strategy 1:} $c_k = \max_{i,j} \frac{\abs{Q_{k,i} - Q_{k,j}}}{d_{\mathcal{X}}\br{i,j}/K}$, $\forall{k \in \bs{K}}$, \emph{i.e.}, we share the privacy budget equally ($\frac{d_{\mathcal{X}}\br{i,j}}{K}$) between the queries. 
	\item \emph{Strategy 2:} $c_k = \max_{i,j} \frac{\norm{Q_{:,i} - Q_{:,j}}_1}{d_{\mathcal{X}}\br{i,j}}$, $\forall{k \in \bs{K}}$, \emph{i.e.}, we add same scale noise to all the query response components. 
	\item \emph{Strategy 3:} We obtain $c$ via Algorithm~\ref{algo:psa}, which distributes the budget between queries proportional to their privacy budget requirements.
\end{enumerate}

For a given multi-linear query $Q \in \mathbb{R}^{K \times N}$, the \emph{improvement factor} of the $d_{\mathcal{X}}$-private Laplace mechanism \eqref{weight-laplace-multi} compared to the baseline ($\epsilon$-differentially private Laplace, \eqref{multi-lap-eps-diff-eq}) mechanism is defined as $\mathrm{IF}\br{Q} := \bc{\frac{\Delta^Q_1 / \epsilon}{c_1}\cdot \frac{\Delta^Q_1 / \epsilon}{c_2} \cdots \frac{\Delta^Q_1 / \epsilon}{c_K}}^{1/K}$, i.e., as a geometric mean of the individual improvement factors. For each $K \in [10]$, we randomly draw $100$ query matrices $Q \in \mathbb{R}^{K \times N}$, and compute the (averaged) improvement factor $\mathrm{IF}\br{Q}$ for the above three different choices of $c$. The results are shown in Figure~\ref{fig:syn-multi-improvement-uniform} and ~\ref{fig:syn-multi-improvement-01}. Some interesting insights are in order. 
\begin{itemize}
    \item Strategy 3 outperforms other strategies for both types of query matrices. This is understandable, as this strategy uses a smarter way of allocating budget between queries. More significantly, the strategy performs much better for the query matrix with binary coefficients (cf. Figure~\ref{fig:syn-multi-improvement-01}). This is true since there is a high likelihood that two query coefficients are the same (i.e., $q_i = q_j$), resulting in no depletion of the privacy budget $d_{\mathcal{X}}(i,j)$.
    \item Strategy 1 has only marginal gain ($\text{IF}(Q) \le 1.5)$ for binary coefficient query matrices. This is because $|Q_{k, i} - Q_{k, j}| \le 1$ for such matrices and therefore the noise scale is essentially $c_k = K/\min_{i, j}(d_{\mathcal{X}}\br{i,j})$, when the query coefficients do not cancel each other out, i.e., $|Q_{k, i} - Q_{k, j}| \neq 0$. This is the same scale as the vanilla Laplace mechanism. The slight improvement is due to the cases where $|Q_{k, i} - Q_{k, j}| = 0$, which does not result in budget depletion in the case of $d_\mathcal{X}$-privacy.
\end{itemize}

\subsection{Single Linear Queries over Real Data}
\label{sec:real-experiments}

\begin{figure*}[t]
	\begin{center}
		\captionsetup[subfigure]{font=scriptsize,labelfont=scriptsize}
		\begin{subfigure}[t]{0.48\textwidth}
			\centering
			\includegraphics[width=0.9\textwidth]{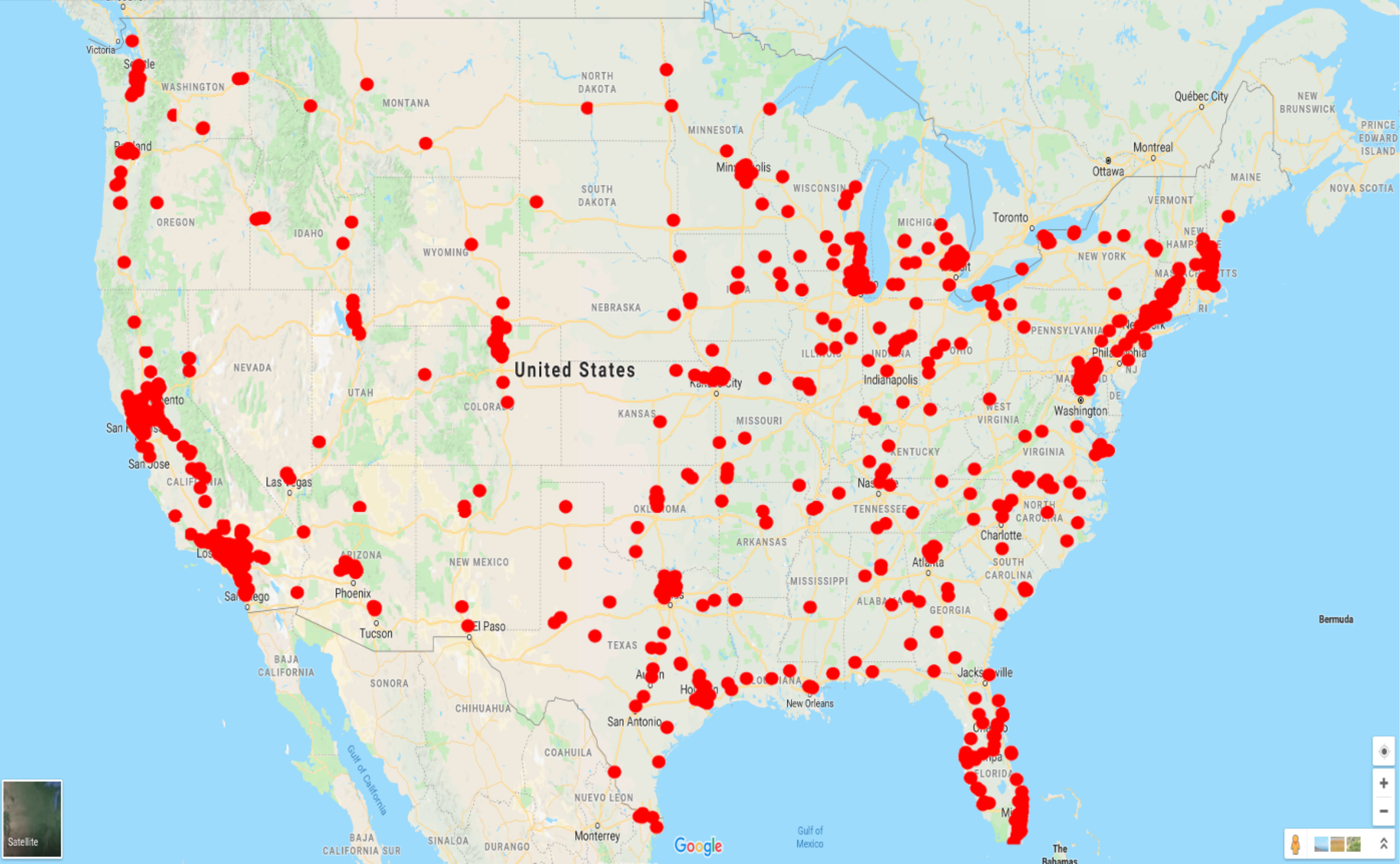}
			\caption{US cities with more than 50k inhabitants}
			\label{fig:real-single-data}
		\end{subfigure}
		\begin{subfigure}[t]{0.48\textwidth}
			\centering
			\includegraphics[width=0.9\textwidth]{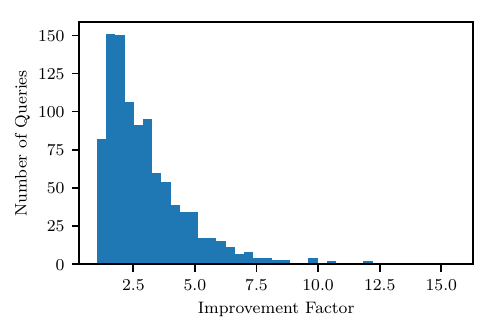}
			\caption{Histogram of the improvement factor (for 1000 random single linear queries)}
			\label{fig:real-single-improvement}
		\end{subfigure}
		\caption{Real-data (US cities \cite{USA-cities}) experiment with $N=741$, and $d_{\mathcal{X}}$ metric defined based on Euclidean distance.}
		\label{fig:real-single}
	\end{center}
\end{figure*}

\begin{figure*}[ht]
	\begin{center}
		\captionsetup[subfigure]{font=scriptsize,labelfont=scriptsize}
		\begin{subfigure}[b]{0.49\textwidth}
			\centering
			\includegraphics[width=1\textwidth]{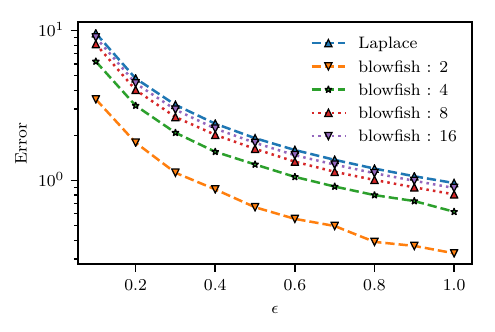}
			\caption{Error under $d_{\mathcal{X}}^{\mathrm{Blow}}$-privacy}
			\label{fig:real-single-data-err}
		\end{subfigure}
		\begin{subfigure}[b]{0.49\textwidth}
			\centering
			\includegraphics[width=1\textwidth]{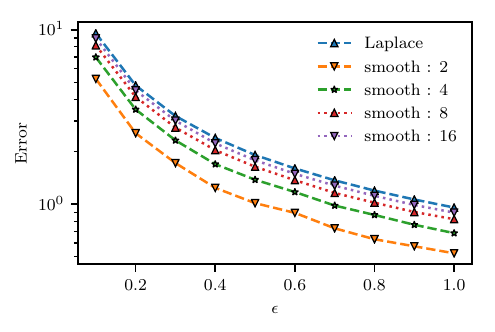}
			\caption{Error under $d_{\mathcal{X}}^{\mathrm{Smooth}}$-privacy}
			\label{fig:real-single-improvement-err}
		\end{subfigure}
		
		\caption{Average RMSE (over 1000 random single linear queries) under different privacy metrics (with $N=50$, and $T = 2,4,8,16$).}
		\label{fig:smooth-blowfish-err}
	\end{center}
\end{figure*}

\begin{figure*}[ht]
	\begin{center}
		\captionsetup[subfigure]{font=scriptsize,labelfont=scriptsize}
		\begin{subfigure}[b]{0.49\textwidth}
			\centering
			\includegraphics[width=1\textwidth]{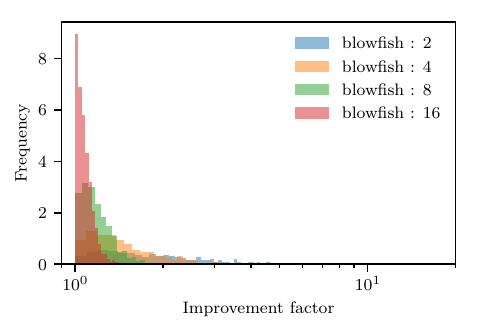}
			\caption{Distribution of the Improvement Factor under $d_{\mathcal{X}}^{\mathrm{Blow}}$-privacy}
			\label{fig:real-single-data-if}
		\end{subfigure}
		\begin{subfigure}[b]{0.49\textwidth}
			\centering
			\includegraphics[width=1\textwidth]{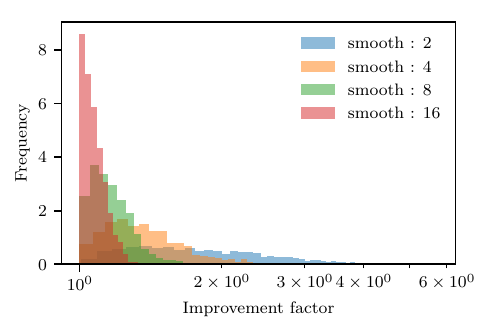}
			\caption{Distribution of the Improvement Factor under $d_{\mathcal{X}}^{\mathrm{Smooth}}$-privacy}
			\label{fig:real-single-improvement-if}
		\end{subfigure}
		
		\caption{Distribution of the improvement factors (for 1000 random single linear queries) under different privacy metrics (with $N=50$, $\epsilon = 1$, and $T = 2,4,8,16$).}
		\label{fig:smooth-blowfish-if}
	\end{center}
\end{figure*}

Next we empirically evaluate $d_{\mathcal{X}}$-private Laplace mechanism  for random single linear queries on a real-world geolocation dataset with longitude, latitude, and elevation attributes. The dataset is based on the United States Cities Database~\cite{USA-cities} which, among other attributes, contains the location (latitude and longitude) and population count of the cities in the United States (US). From this dataset, we extract the location and population count of cities with more than 50k inhabitants, resulting in a total of 741 cities. We further augment this dataset with elevation information by querying the Google Maps Elevation API~\cite{elevation-API} with the corresponding latitude and longitude values. We translate this dataset into a histogram over the cities (with $N = 741$). The 2D-locations (longitude and latitude wise) of the towns are presented in Figure~\ref{fig:real-single-data}. We define the privacy budget $d_{\mathcal{X}}$ based on the Euclidean distance on this 2D-representation. 

We generate and evaluate $1000$ random linear queries over this dataset. The improvement factors of these queries are presented in Figure~\ref{fig:real-single-improvement}. The average of the IF values lies between 2 to 3, with some queries showing an improvement factor of more than 7.5. We also wanted to test the improvement factor over queries with an obvious real-world interpretation. One such query is the ``average elevation of a US resident's house'' (the coefficients are simply the elevation of each city). In this case our algorithm performed particularly well, with an improvement factor of $202$. This scale of improvement is due to the fact that there is a strong correlation between the query and the distance map: two nearby cities (\emph{i.e.,} having strong privacy requirement) also have similar elevation. Note that even if the solutions shown above are sub-optimal, we still perform better than the baseline for synthetic data, and outperform it depending on the query structure and real data.  

\subsection{Experiments with Blowfish Privacy}
\label{sec:blow-experiments}
In this section, we demonstrate that the $d_\mathcal{X}$-privacy notion can generalize some of the other alternative privacy notions as well, and hence our techniques can be applied to these other notions. This is true since our general pre-processing strategy applies to \emph{any metric}. Thus, for instance, our techniques can be extended to the Blowfish \cite{he2014blowfish} privacy (without constraints) notion as well. We can carefully define a $d_{\mathcal{X}}$ metric for any privacy policy considered in the Blowfish framework. First, we define $d_{\mathcal{X}}$ such that $d_{\mathcal{X}} \br{i,j} = \infty , \forall{i \neq j}$, and $d_{\mathcal{X}} \br{k,k} = 0 , \forall{k}$. Then for each pair of neighbors $\br{i,j}$, we check if there is a secret to be protected with the Blowfish policy. If so we just set $d_{\mathcal{X}} \br{i,j} = \epsilon$ (the privacy budget). Finally, we need to make sure (possibly by some transformations) that the resulting $d_{\mathcal{X}}$ satisfies the triangular inequality (a necessary condition for a distance metric).

We consider the same data, single linear queries, mechanism ($d_\mathcal{X}$-private Laplace) and performance measure (squared loss) used in Section~\ref{sec:syn-experiments-single}. But here we work with two different privacy metrics. Given a threshold $T$, and a privacy parameter $\epsilon$, define:
\begin{enumerate}
\item $d_{\mathcal{X}}^{\mathrm{Blow}}$ s.t. $d_{\mathcal{X}}^{\mathrm{Blow}}\br{i,j} = \epsilon$ if $d_{\mathcal{X}}^{\mathrm{Euc}}\br{i,j} \leq T, \text{ and}\\ d_{\mathcal{X}}^{\mathrm{Blow}}\br{i,j} = \infty$ otherwise
\item $d_{\mathcal{X}}^{\mathrm{Smooth}}$ s.t. $d_{\mathcal{X}}^{\mathrm{Smooth}}\br{i,j} = \epsilon$ if $d_{\mathcal{X}}^{\mathrm{Euc}}\br{i,j} \leq T, \text{ and}\\ d_{\mathcal{X}}^{\mathrm{Smooth}}\br{i,j} = \frac{\epsilon d_{\mathcal{X}}^{\mathrm{Euc}}\br{i,j}}{T}$ otherwise,
\end{enumerate}
where $d_{\mathcal{X}}^{\mathrm{Euc}}\br{i,j} := \sqrt{\br{u_i - u_j}^2 + \br{v_i - v_j}^2}$.

The first metric assigns privacy budget $\epsilon$ for any pair of points within distance $T$, and $\infty$ otherwise. The second metric ``smoothly'' increases the privacy budget proportional to the distance between the pair of points. Our base method for comparison is the $\epsilon$-differentially private Laplace mechanism. First, we compute the average RMSE over $1000$ random single linear queries under both privacy metrics defined above (for different values of $\epsilon$ and $T$). The results are shown in Figure~\ref{fig:smooth-blowfish-err}. We can see that the results under both metrics are roughly the same. The higher the threshold $T$ (\emph{i.e.}, more neighbors are protected), the higher is the average error. The $d_{\mathcal{X}}^{\mathrm{Smooth}}$ metric behaves like the $d_{\mathcal{X}}^{\mathrm{Euc}}$ metric after the threshold value. Thus it induces tighter (and smoother) privacy than $d_{\mathcal{X}}^{\mathrm{Blow}}$, and results in a higher average error for the same threshold.

Then we fix $\epsilon = 1$, and for each random query ($1000$ in total), we compute the improvement factor. The resulted values are presented in a distribution form in Figure~\ref{fig:smooth-blowfish-if}. Observe that for higher threshold values the distributions under both metrics are roughly similar, but for lower threshold values (\emph{e.g.} $T = 2$), the improvement factor under $d_{\mathcal{X}}^{\mathrm{Blow}}$ is better than under $d_{\mathcal{X}}^{\mathrm{Smooth}}$.



\section{Discussion}
\label{sec:discuss}
\descr{Example $d_\mathcal{X}$-metric Instantiations:}
The main contribution of this paper is a meta procedure that converts an existing differentially private mechanism to its $d_\mathcal{X}$-private counterpart, given any metric $d_\mathcal{X}$. 
The interpretation of the privacy guarantees of the resulting mechanism is tied to how well the metric translates a given set of privacy requirements. Here we show some examples of appropriate $d_\mathcal{X}$-metrics for different privacy requirements.

\textit{Location Privacy:} We have already presented some location privacy specific $d_\mathcal{X}$-metrics, i.e., 
the Euclidean distance based metric in Section~\ref{sec:syn-experiments-single} where nearby points are required to be more indistinguishable than distant points, and the distance threshold metrics in Section~\ref{sec:blow-experiments} (based on an example of sensitive information specification for the Blowfish framework~\cite{he2014blowfish}), which provides higher indistinguishability for points that are within a given distance threshold. 

\textit{Heterogeneous Privacy for Tabular Data:} Notably, location privacy is not the only application for $d_\mathcal{X}$-privacy. We have shown one such instance in Example~\ref{ex:example} where the metric defines some attribute values as more sensitive than others. First, for binary datasets (each attribute having a cardinality of two), the metric in Example~\ref{ex:example} can be  generalized for any number of attributes. This does not generalize to attributes with more than 2 values, as the $\min$ function used in the metric does not satisfy the triangle inequality in such a case. An example metric, for the same privacy requirement (i.e., providing higher privacy to individuals having selected attribute values), can be defined as:
\[
d_{\mathcal{X}}(i, j) = \sum_{k = 1}^d \left( \epsilon(X_i^{(k)}) + \epsilon(X_j^{(k)}) \right) \ind{X_i^{(k)} \neq X_j^{(k)}}.
\]
where $d$ is the total number of attributes in the dataset and $\epsilon(X_i^{(k)})$ defines the privacy budget allocated to the $k$th attribute value. One can then set $\epsilon(X_i^{(k)}) = \epsilon_k$ for all $i \in [N]$, where $\epsilon_k$ can be set to be lower for more sensitive attribute values. This closely resembles the metric $d_{\mathcal{X}}(i, j) = \sum_{k = 1}^d \epsilon_k \ind{X_i^{(k)} \neq X_j^{(k)}}$ (discussed in the introduction) which allows to set sensitivity of the entire attribute via assigning the same $\epsilon_k$ for all values of the attribute. Note that this privacy metric relates to the notion of heterogeneous differential privacy~\cite{alaggan2015heterogeneous} in which a user (owner of $d$ items) chooses a separate privacy budget $\epsilon_k$ for its $k$th item. Similarly, if we require more privacy for some individuals in the dataset (modelled as elements in the histogram representation of the dataset), we can use the metric $d_{\mathcal{X}}(i, j) = (\epsilon_i + \epsilon_j) \ind{i \neq j}$, and assign lower privacy budgets for more sensitive elements. 

\textit{Data Generalization:} Several other examples of $d_\mathcal{X}$-metrics are given in~\cite{chatzikokolakis2013broadening}. One example is when the exact date of a particular event is considered sensitive, but releasing a slightly generalized date, say within a $T$-day period, might be appropriate. In this case the scaled metric $d_\mathcal{X}(i, j) = \epsilon \frac{| u_i - u_j |}{T}$ can be used, where $u_i$ is the exact date (say, number of days since January 1, 2000) associated with the $i$th element in the data universe~\cite{chatzikokolakis2013broadening}. 

\textit{Privacy of Time-Series Data:} Another example is protecting time-series data (e.g., smart energy data) where the privacy requirement is to only prevent fine-grained inference of the time-series. Here, an $l_\infty$ norm based $d_\mathcal{X}$-metric is appropriate which is the maximum of the distances between each component of the time-series~\cite{chatzikokolakis2013broadening}. 

\textit{Privacy in Social Networks:}
Another natural metric based on a minimum spanning tree is given in~\cite{he2014blowfish}: vertices represent elements of the data universe, with edges between them having equal weights. Here the adversary may better distinguish points farther apart in the tree, than those that are closer. If some elements of the data universe are highly sensitive than others, non-uniform edge weights can capture the requirement. This metric is suitable for privacy in social networks. 

A comprehensive treatment of privacy requirements and a suitable choice of $d_\mathcal{X}$-metric for each of them is beyond the scope of this work. The above examples show that $d_\mathcal{X}$-privacy can be used in many different applications. We stress however that the metric $d_\mathcal{X}$ must be appropriately defined to achieve meaningful privacy goals. A wrong choice of $d_\mathcal{X}$-metric may adversely impact privacy. For instance, if we replace the $\min$ function with the $\max$ function in Eq.~\ref{eq:ex-dx} of Example~\ref{ex:example}, then even though the resulting function is still a metric, it does not satisfy the privacy requirement of providing more protection to more sensitive attributes. In particular, the query $q = ({\tt{MNA}}, {\tt{MNB}}, {\tt{FNA}}, {\tt{FNB}}) = ({\tt{N}})$ will now be answered with noise of scale $c = 1/\epsilon_1$. 

\descr{Correlated Data:} If the database contains correlated data, it may be possible to infer about sensitive attributes even if the mechanism is $d_\mathcal{X}$-private. For instance, in Example~\ref{ex:example}, it may be known that males above 18 years of age are 90 percent more likely to be native. Then the answer to the query $(\tt{MA}) = (\tt{MYA}, \tt{MNA})$ will have less noise added to it as the query is non-sensitive (noise scale will be $c = 1/\epsilon_1$ in the example). Multiplying the answer by $0.9$ gives us a much more accurate approximate number of native men above the age of 18, then what would have been possible through the query  $(\tt{MYA})$ (noise scale $c = 1/\epsilon_0$). Protecting the answers from such correlations requires broadening the scope of $d_\mathcal{X}$-privacy to take such information as input, possibly in the form of constraints, as is done in the Blowfish privacy framework~\cite{he2014blowfish}. We note that susceptibility of $d_\mathcal{X}$-privacy under these \emph{column-wise} correlations is similar to the case of differential privacy with correlated rows. Just like how differential privacy provides privacy for atypical rows (uncorrelated rows), $d_\mathcal{X}$-privacy guarantees privacy for atypical attribute values.  



\descr{Unbounded Differential Privacy:} In many instantiations of our meta procedure \eqref{abstratc-pre-opt}, we have used the bounded differential privacy model (in which the number of elements, i.e., $n$, in the dataset is public information). Our procedure can also be applied to unbounded differential privacy by spending some privacy budget to query database size $n$, similar to the conversion between the two flavours of differential privacy~\cite[p. 358]{salil-tut}. For the Laplace mechanism \eqref{opt-laplace-multi-linear-query}, by setting $cq’ = q$, we could get rid of the dependence on $n$, which makes it applicable to unbounded differential privacy as well. For SmallDB/MWEM, it's better to exploit the knowledge $n$, as we need to do the pre-processing only once for a given query set $Q$ and dataset $x$.


\section{Related Work}
\label{sec:related-work}
In \cite{andres2013geo} the notion of geo-indistinguishability is proposed which protects a user's exact location while allowing approximate information for location-based services. Some mechanisms to achieve privacy under this notion are also proposed which are variations of the Laplace mechanism for differential privacy. Geo-indistinguishability can be considered as an example of $d_\mathcal{X}$-privacy where the Euclidean metric within the discrete Cartesian plane is used as the data universe. Compared to \cite{andres2013geo}, where only a few variations of the Laplace mechanism are given, we have proposed a general procedure to convert any differential privacy mechanism to its $d_\mathcal{X}$-privacy equivalent for linear queries. Furthermore, the focus of~\cite{andres2013geo} is on location based services in the local model, whereas our work targets $d_\mathcal{X}$-private mechanisms for linear queries over histograms in the centralized model.


As mentioned earlier, the definition of $d_\mathcal{X}$-privacy is an instance of the notion of \emph{generalized privacy} with a metric $d_\mathcal{X}$ which was proposed in~\cite{chatzikokolakis2013broadening} for the case of statistical databases (where each user's data is one row of the database).  
In addition to proposing the definition, the authors in \cite{chatzikokolakis2013broadening} have only constructed \textit{universally optimal mechanisms} \cite{ghosh2012universally}\footnote{Roughly, a mechanism is universally optimal if it provides the same utility to all users, regardless of their background information and (legal) loss function (modeling utility loss), as would a mechanism that is specifically tailored to each user.} under some specific $d_\mathcal{X}$ metrics (such as Manhattan metric) for some particular class of queries such as count, sum, average, and percentage queries. In comparison, we propose a generic strategy to tailor any differentially private mechanism to satisfy $d_\mathcal{X}$-privacy for linear queries (which encompass a broad range of queries including the aforementioned). 

Blowfish privacy~\cite{he2014blowfish} is a class of definitions that aims to strengthen differential privacy by the use of privacy policies that include a set of secrets (i.e., information deemed sensitive in the dataset, akin to what is modelled by the $d_\mathcal{X}$-metric) and a set of constraints that model an adversary's background knowledge or public knowledge about the dataset. There are some recent results on generalizing differentially private mechanisms to the Blowfish privacy equivalent under a given privacy policy~\cite{haney-blowfish}. In contrast to~\cite{haney-blowfish}, we (a) consider any instance of the $d_{\mathcal{X}}$-metric (which covers Blowfish \cite{he2014blowfish} privacy notion without constraints), and (b) pre-process the query alone (and not the input database) -- thus we only need to do pre-processing once for a given data domain, i.e., not having to redo pre-processing for database changes. Currently, our proposed procedure applies to only a special case of the Blowfish that does not introduce deterministic constraints (modelling public knowledge), and extending our results to general Blowfish which deals with correlations is an interesting future direction. 

The concept of \emph{heterogeneous differential privacy} is proposed in \cite{alaggan2015heterogeneous} in the the user profile setting where a database itself is attributed to a single user. They consider the case where a user does not have homogeneous privacy requirements for all his/her items. We note that our meta procedure idea can be extended to the user profile setting as well. This extension would require slight modification in privacy and sensitivity definitions and utility analysis. In particular, the metric $d_{\mathcal{X}}\br{u,v} = \sum_{i=1}^{d}{\epsilon_i \ind{u_i \neq v_i}}$ (that we discussed in the introduction and at the end of Appendix~\ref{sec:stat-query}) is closely related to the privacy definition in \cite{alaggan2015heterogeneous}. For linear queries, the stretching mechanism in \cite{alaggan2015heterogeneous} also transforms the original query vector $q$ into $q’$ (similar to our meta procedure), but their noise term is fixed and depends on the {\em global} sensitivity (in that sense our meta procedure is more general than theirs with noise parameter $c$). Moreover, the transformation $q \mapsto q’$ in \cite{alaggan2015heterogeneous} is not utility dependent. In the context of user profiles, detailed investigation of the connection between our (extended) meta procedure and the stretching mechanism is indeed an interesting future work. Similarly, \cite{jorgensen2015conservative} considers \emph{personalized differential privacy} (PDP) where different users have different privacy expectations in the context of statistical databases. However, unlike $d_\mathcal{X}$-privacy, a user can only set the same privacy budget for all items (column-wise privacy). Moreover, a general \emph{sampling} mechanism to convert any differential privacy mechanism to its PDP counterpart is also proposed in ~\cite{jorgensen2015conservative}, which samples rows from the original dataset based on the privacy requirement of each user. This introduces an additional error term (due to sampling)~\cite{jorgensen2015conservative}. In both these prior works, the utility measure of interest is not taken into consideration while distributing the privacy budget, whereas our meta procedure explicitly focuses on the utility measure.



\section{Conclusion}
\label{sec:conclude}

In this paper, we developed new $d_{\mathcal{X}}$-private mechanisms for linear queries by extending the standard $\epsilon$-differentially private mechanisms. These new mechanisms fully utilize the privacy budgets of different elements and maximize the utility of the private response. We have empirically shown that carefully selecting the model parameters of the $d_{\mathcal{X}}$-private mechanisms (depending on the utility function and $d_{\mathcal{X}}$-metric) can result in substantial improvement over the baseline mechanisms in terms of utility. Note that our analysis can be extended to advanced $\epsilon$-differentially private mechanisms such as the Matrix~\cite{li2011efficient}, and $K$-norm~\cite{hardt2010geometry} mechanisms. We leave it as future work. Finally, we would like to remark that for statistical queries (a special case of linear queries), which are (loosely) defined as the sum of predicates over the rows of the input dataset, we can design $d_{\mathcal{X}}$-private mechanisms more efficiently by exploiting the sum-structure. We refer the reader to Appendix~\ref{sec:stat-query} for more details.

\section{Acknowledgment}
\label{sec:ack}
When the work was done, both Parameswaran Kamalaruban, and Victor Perrier were working at Data61, CSIRO.

\newpage

\appendix
\section{Proofs}
\label{sec:proofs}

\subsection{Laplace Mechanism}
\label{app:lap-mech}
\begin{reptheorem}{laplace-privacy-prop}
	\label{repprop:laplace-privacy-prop}
	If $\Delta_1^{q'} \br{i,j} \leq d_{\mathcal{X}}\br{i,j}$, $\forall{i,j \in \bs{N}}$, then the mechanism $\mathcal{M}_{\mathrm{Lap},c}\br{\cdot , c \odot q'}$ given by \eqref{weight-laplace-multi} satisfies $d_{\mathcal{X}}$-privacy. 
\end{reptheorem}
\begin{proof}
	Let $x,x' \in \mathbb{R}^N$ s.t. $\norm{x-x'}_1 \leq 2$, $x_i \neq x'_i$, and $x_j \neq x'_j$, and let $q \in \mathcal{Q}$. Let $p_x$ and $p_{x'}$ denote the probability density functions of $\mathcal{M}_{\mathrm{Lap},c}\br{x , q}$ and $\mathcal{M}_{\mathrm{Lap},c}\br{x' , q}$ respectively. Then for any $z \in \mathcal{Y}$ we have
	\begin{align*}
	\frac{p_x \br{z}}{p_{x'} \br{z}} ~=~& \Pi_{i=1}^k \br{\frac{\exp\br{-\frac{\abs{c_i q'\br{x}_i - z_i}}{c_i}}}{\exp\br{-\frac{\abs{c_i q'\br{x'}_i - z_i}}{c_i}}}} \\	
	~=~& \Pi_{i=1}^k \exp\br{\frac{\abs{c_i q'\br{x'}_i - z_i} - \abs{c_i q'\br{x}_i - z_i}}{c_i}} \\
	~\overset{(i)}{\leq}~& \Pi_{i=1}^k \exp\br{\frac{\abs{c_i q'\br{x'}_i - c_i q'\br{x}_i}}{c_i}} \\
	~=~& \exp\br{\norm{q'\br{x} - q'\br{x'}}_1} \\
	~\overset{(ii)}{\leq}~& \exp\br{\Delta_1^{q'} \br{i,j}} \\
	~\leq~& \exp\br{d_{\mathcal{X}} \br{i,j}} ,
	\end{align*}
	where $(i)$ follows from the triangle inequality and $(ii)$ follows from the definition of generalized global sensitivity and due to the choice of $x$ and $x'$. That $\frac{p_x \br{z}}{p_{x'} \br{z}} \geq \exp\br{- d_{\mathcal{X}} \br{i,j}}$, follows by symmetry. 
\end{proof} 

\begin{reptheorem}{laplace-utility-prop}
	\label{repprop:laplace-utility-prop}
	Let $Q: \mathbb{R}^N \rightarrow \mathbb{R}^k$ be a multi-linear query of the form $Q\br{x} = Q x$, and let $\textnormal{\textsf{Z}} = \mathcal{M}_{\mathrm{Lap},c} \br{x,c\odot Q'} = c\odot Q'x + \textnormal{\textsf{Y}}$ with $\textnormal{\textsf{Y}}_i \overset{\perp}{\sim} \mathrm{Lap}\br{c_i}$. 
	\begin{enumerate}
		\item When $\ell_2^2 \br{y,y'} = \norm{y - y'}_2^2$, we have
		\begin{align*}
		    \mathrm{err}_{\ell_2^2}\br{\mathcal{M}_{\mathrm{Lap},c} \br{\cdot,c\odot Q'},Q} ~\leq~& 2 n^2 \norm{c\odot Q' - Q}_{2}^2 \\
		    & \quad + 4 \norm{c}_2^2 ,
		\end{align*}
		where $\mathrm{err}_{\ell}\br{\mathcal{M},Q}$ is defined in \eqref{error-def-eq}.
		\item When $\ell_p \br{y,y'} = \norm{y - y'}_p$, we have
		\begin{align*}
		    \mathrm{err}_{\ell_p}\br{\mathcal{M}_{\mathrm{Lap},c} \br{\cdot,c\odot Q'},Q} ~\leq~& n \norm{c\odot Q' - Q}_{p}  \\ 
		    & + \Ee{\textnormal{\textsf{Y}}_i \overset{\perp}{\sim} \mathrm{Lap}\br{c_i}}{\norm{\textnormal{\textsf{Y}}}_p} .
		\end{align*}
		Note that $\Ee{\textnormal{\textsf{Y}}_i \overset{\perp}{\sim} \mathrm{Lap}\br{c_i}}{\norm{\textnormal{\textsf{Y}}}_1} ~=~ \norm{c}_1$.
		\item $\forall{\delta \in (0,1]}$, with probability at least $1-\delta$ we have 
		\[
		\norm{Q x - \textnormal{\textsf{Z}}}_\infty  \leq n \norm{c\odot Q' - Q}_{\infty} + \ln\br{\frac{k}{\delta}} \cdot \norm{c}_{\infty} .
		\]
	\end{enumerate}
\end{reptheorem}
\begin{proof}
	\textbf{Part 1.} Consider
	\begin{align*}
	& \Ee{\textsf{Z}}{\ell \br{\textsf{Z},Q\br{x}}} \\
	~=~& \Ee{\textsf{Z}}{\norm{\textsf{Z} - Q x}_2^2} \\ 
	~=~& \Ee{\textsf{Y}_i \overset{\perp}{\sim} \mathrm{Lap}\br{c_i}}{\norm{c \odot Q' x + \textsf{Y} - Q x}_2^2} \\
	~\overset{(i)}{\leq}~& \Ee{\textsf{Y}_i \overset{\perp}{\sim} \mathrm{Lap}\br{c_i}}{\br{\norm{c \odot Q' x - Q x}_2 + \norm{\textsf{Y}}_2}^2} \\
	~\overset{(ii)}{\leq}~& 2 \Ee{\textsf{Y}_i \overset{\perp}{\sim} \mathrm{Lap}\br{c_i}}{\norm{c \odot Q' x - Q x}_2^2 + \norm{\textsf{Y}}_2^2} \\
	~=~& 2 \bc{\norm{c \odot Q' x - Q x}_2^2 + \Ee{\textsf{Y}_i \overset{\perp}{\sim} \mathrm{Lap}\br{c_i}}{\norm{\textsf{Y}}_2^2}} \\
	~=~& 2 \bc{\sum_{i=1}^{k}{\abs{\ip{c_i Q'_{i,:} - Q_{i,:}}{x}}^2} + \Ee{\textsf{Y}_i \overset{\perp}{\sim} \mathrm{Lap}\br{c_i}}{\norm{\textsf{Y}}_2^2}} \\
	~\overset{(iii)}{\leq}~& 2 \bc{\sum_{i=1}^{k}{\norm{c_i Q'_{i,:} - Q_{i,:}}_2^2\norm{x}_2^2} + \Ee{\textsf{Y}_i \overset{\perp}{\sim} \mathrm{Lap}\br{c_i}}{\norm{\textsf{Y}}_2^2}} \\
	~=~& 2 \bc{\norm{x}_2^2 \sum_{i=1}^{k}{\norm{c_i Q'_{i,:} - Q_{i,:}}_2^2} + \Ee{\textsf{Y}_i \overset{\perp}{\sim} \mathrm{Lap}\br{c_i}}{\norm{\textsf{Y}}_2^2}} \\
	~\overset{(iv)}{\leq}~& 2 \bc{n^2 \sum_{i=1}^{k}{\norm{c_i Q'_{i,:} - Q_{i,:}}_2^2} + \Ee{\textsf{Y}_i \overset{\perp}{\sim} \mathrm{Lap}\br{c_i}}{\norm{\textsf{Y}}_2^2}} \\
	~=~& 2 \bc{n^2 \norm{c \odot Q' - Q}_{2}^2 + \Ee{\textsf{Y}_i \overset{\perp}{\sim} \mathrm{Lap}\br{c_i}}{\norm{\textsf{Y}}_2^2}} \\
	~\overset{(v)}{=}~& 2 \bc{n^2 \norm{c \odot Q' - Q}_{2}^2 + 2 \norm{c}_2^2} 
	\end{align*}
	where $(i)$ is by triangle inequality, $(ii)$ is due to the fact that $(a+b)^2 \leq 2a^2 + 2b^2$, $(iii)$ is by H\"older's Inequality, $(iv)$ is due to the fact that $\norm{x}_2 \leq \norm{x}_1 = n$, and $(v)$ is due to the fact that $\Ee{\textsf{Y}_i \overset{\perp}{\sim} \mathrm{Lap}\br{c_i}}{\norm{\textsf{Y}}_2^2} = 2 \norm{c}_2^2$ (since $\Ee{\textsf{X} \sim \mathrm{Lap}\br{\lambda}}{\textsf{X}^2} = 2 \lambda^2$ for $\textsf{X} \in \mathbb{R}$). This completes the proof of first part. 
	
	\textbf{Part 2.} Consider (by the similar reasoning as of Part 1)
	\begin{align*}
	& \Ee{\textsf{Z}}{\ell \br{\textsf{Z},Q\br{x}}} \\
	~=~& \Ee{\textsf{Z}}{\norm{\textsf{Z} - Q x}_p} \\ 
	~=~& \Ee{\textsf{Y}_i \overset{\perp}{\sim} \mathrm{Lap}\br{c_i}}{\norm{c \odot Q' x + \textsf{Y} - Q x}_p} \\
	~\leq~& \Ee{\textsf{Y}_i \overset{\perp}{\sim} \mathrm{Lap}\br{c_i}}{\norm{c \odot Q' x - Q x}_p + \norm{\textsf{Y}}_p} \\
	~=~& \norm{c \odot Q' x - Q x}_p + \Ee{\textsf{Y}_i \overset{\perp}{\sim} \mathrm{Lap}\br{c_i}}{\norm{\textsf{Y}}_p} \\ 
	~=~& \norm{c \odot Q' x - Q x}_p + \Ee{\textsf{Y}_i \overset{\perp}{\sim} \mathrm{Lap}\br{c_i}}{\norm{\textsf{Y}}_p} \\ 
	~=~& \br{\sum_{i=1}^{k}{\abs{\ip{c_i Q'_{i,:} - Q_{i,:}}{x}}^p}}^{1/p} + \Ee{\textsf{Y}_i \overset{\perp}{\sim} \mathrm{Lap}\br{c_i}}{\norm{\textsf{Y}}_p} \\ 
	~\leq~& \br{\sum_{i=1}^{k}{\norm{c_i Q'_{i,:} - Q_{i,:}}_p^p\norm{x}_q^p}}^{1/p} + \Ee{\textsf{Y}_i \overset{\perp}{\sim} \mathrm{Lap}\br{c_i}}{\norm{\textsf{Y}}_p} \\  
	~=~& \norm{x}_q \br{\sum_{i=1}^{k}{\norm{c_i Q'_{i,:} - Q_{i,:}}_p^p}}^{1/p}  + \Ee{\textsf{Y}_i \overset{\perp}{\sim} \mathrm{Lap}\br{c_i}}{\norm{\textsf{Y}}_p} \\ 
	~\leq~& n \br{\sum_{i=1}^{k}{\norm{c_i Q'_{i,:} - Q_{i,:}}_p^p}}^{1/p} + \Ee{\textsf{Y}_i \overset{\perp}{\sim} \mathrm{Lap}\br{c_i}}{\norm{\textsf{Y}}_p} \\ 
	~=~& n \norm{c \odot Q' - Q}_{p} + \Ee{\textsf{Y}_i \overset{\perp}{\sim} \mathrm{Lap}\br{c_i}}{\norm{\textsf{Y}}_p} .
	\end{align*}
	Note that $\Ee{\textsf{Y}_i \overset{\perp}{\sim} \mathrm{Lap}\br{c_i}}{\norm{\textsf{Y}}_1} = \norm{c}_1$ (since $\Ee{\textsf{X} \sim \mathrm{Lap}\br{\lambda}}{\abs{\textsf{X}}} = \lambda$ for $\textsf{X} \in \mathbb{R}$).
	
	\textbf{Part 3.} We will use the fact that if $\textsf{Y} \sim \mathrm{Lap}\br{b}$, then $\P{\abs{\textsf{Y}} \geq t \cdot b} = \exp\br{-t}$. We have:
	\begin{align*}
	& \P{\norm{c \odot Q' x - \textsf{Z}}_\infty  \geq \ln\br{\frac{k}{\delta}} \cdot \norm{c}_{\infty}} \\
	~=~& \P{\max_{i \in \bs{k}}{\abs{\textsf{Y}_i}}  \geq \ln\br{\frac{k}{\delta}} \cdot \norm{c}_{\infty}} \\
	~\leq~& k \cdot \P{\abs{\textsf{Y}_i}  \geq \ln\br{\frac{k}{\delta}} \cdot \norm{c}_{\infty}} \\
	~\leq~& k \cdot \P{\abs{\textsf{Y}_i}  \geq \ln\br{\frac{k}{\delta}} \cdot c_i} \\
	~=~& k \cdot \br{\frac{\delta}{k}} \\
	~=~& \delta
	\end{align*}
	where the first inequality is due to union bound, and the second to last equality follows from the fact that each $\textsf{Y}_i \sim \mathrm{Lap}\br{c_i}$. That is with probability at least $1-\delta$ we have 
	\begin{align*}
	& \norm{Q x - \textsf{Z}}_\infty  \\
	~\leq~& \norm{c \odot Q' x - Q x}_\infty + \ln\br{\frac{k}{\delta}} \cdot \norm{c}_{\infty} \\
	~=~& \max_{i \in \bs{k}}{\abs{\ip{c_i Q'_{i,:} - Q_{i,:}}{x}}} + \ln\br{\frac{k}{\delta}} \cdot \norm{c}_{\infty} \\
	~\leq~& \max_{i \in \bs{k}}{\norm{c_i Q'_{i,:} - Q_{i,:}}_\infty \norm{x}_1} + \ln\br{\frac{k}{\delta}} \cdot \norm{c}_{\infty} \\
	~=~& n \max_{i \in \bs{k}}{\norm{c_i Q'_{i,:} - Q_{i,:}}_\infty} + \ln\br{\frac{k}{\delta}} \cdot \norm{c}_{\infty} \\
	~=~& n \norm{c \odot Q' - Q}_{\infty} + \ln\br{\frac{k}{\delta}} \cdot \norm{c}_{\infty} .
	\end{align*}
\end{proof}

\subsection{Exponential Mechanism}
\label{app:exp-mech}
\begin{reptheorem}{exp-privacy-theorem}
	\label{repthm:exp-privacy-theorem}
	If $\Delta u' \br{i,j} \leq cd_{\mathcal{X}}\br{i,j}, \forall{i,j \in \bs{N}}$, then the mechanism $\mathcal{M}_{\mathrm{Exp},c}\br{\cdot,u'}$ satisfies the $d_{\mathcal{X}}$-privacy.
\end{reptheorem}
\begin{proof}
	For clarity, we assume $\mathcal{R}$ to be finite. Let $x,x' \in \mathbb{R}^N$ s.t. $\norm{x-x'}_1 \leq 2$, $x_i \neq x'_i$ and $x_j \neq x'_j$. Then for any $r \in \mathcal{R}$ we have
	\begin{align*}
	& \frac{\P{\mathcal{M}_{\mathrm{Exp},c}\br{x , u'} = r}}{\P{\mathcal{M}_{\mathrm{Exp},c}\br{x' , u'} = r}} \\
	~=~& \frac{\br{\frac{\exp\br{\frac{u'\br{x,r}}{2 c}}}{\sum_{r' \in \mathcal{R}}{\exp\br{\frac{u'\br{x,r'}}{2 c}}}}}}{\br{\frac{\exp\br{\frac{u'\br{x',r}}{2 c}}}{\sum_{r' \in \mathcal{R}}{\exp\br{\frac{u'\br{x',r'}}{2 c}}}}}} \\	
	~=~& \frac{\exp\br{\frac{u'\br{x,r}}{2 c}}}{\exp\br{\frac{u'\br{x',r}}{2 c}}} \cdot \frac{\sum_{r' \in \mathcal{R}}{\exp\br{\frac{u'\br{x',r'}}{2 c}}}}{\sum_{r' \in \mathcal{R}}{\exp\br{\frac{u'\br{x,r'}}{2 c}}}} \\
	~=~& \exp\br{\frac{u'\br{x,r} - u'\br{x',r}}{2 c}} \cdot \frac{\sum_{r' \in \mathcal{R}}{\exp\br{\frac{u'\br{x',r'}}{2 c}}}}{\sum_{r' \in \mathcal{R}}{\exp\br{\frac{u'\br{x,r'}}{2 c}}}} \\
	~\leq~& \exp\br{\frac{\Delta u' \br{i,j}}{2 c}} \cdot \frac{\sum_{r' \in \mathcal{R}}{\exp\br{\frac{u'\br{x,r'} + \Delta u' \br{i,j}}{2 c}}}}{\sum_{r' \in \mathcal{R}}{\exp\br{\frac{u'\br{x,r'}}{2 c}}}} \\
	~=~& \exp\br{\frac{\Delta u' \br{i,j}}{2 c}} \cdot \exp\br{\frac{\Delta u' \br{i,j}}{2 c}} \cdot \br{1} \\
	~=~& \exp\br{d_{\mathcal{X}} \br{i,j}} .
	\end{align*}
	Similarly, $\frac{\P{\mathcal{M}_{\mathrm{Exp},c}\br{x , u'} = r}}{\P{\mathcal{M}_{\mathrm{Exp},c}\br{x' , u'} = r}} \geq \exp\br{- d_{\mathcal{X}} \br{i,j}}$ by symmetry. 
\end{proof}

\begin{reptheorem}{exp-utility-theorem}
	\label{repthm:exp-utility-theorem}
	Fixing a database $x$, let 
	\[
	\mathcal{R}_{\star_{u'}} = \bc{r \in \mathcal{R} : u'\br{x,r} = \star_{u'} \br{x}}
	\]
	denote the set of elements in $\mathcal{R}$ which attain utility score $\star_{u'} \br{x}$. Also define $\delta_{u,u'} := \max_{x,r}{\abs{u\br{x,r} - u'\br{x,r}}}$. Then for $\textnormal{\textsf{Z}} = \mathcal{M}_{\mathrm{Exp},c}\br{x , u'}$, we have
	\begin{align*}
	& \P{u \br{x, \textnormal{\textsf{Z}}} \leq \delta_{u,u'} + \star_{u'} \br{x} - 2 c\bc{\ln\br{\frac{\abs{\mathcal{R}}}{\abs{\mathcal{R}_{\star_{u'}}}}} + t}} \\
	\leq~& e^{-t} . 
	\end{align*}
	Since we always have $\abs{\mathcal{R}_{\star_{u'}}} \geq 1$, we get
	\[
	\P{u \br{x, \textnormal{\textsf{Z}}} \leq \delta_{u,u'} + \star_{u'} \br{x} - 2 c\bc{\ln\br{\abs{\mathcal{R}}} + t}} \leq e^{-t} . 
	\]
\end{reptheorem}
\begin{proof}
	\begin{align*}
	\P{u' \br{x, \textsf{Z}} \leq \alpha} ~\leq~& \frac{\abs{\mathcal{R}} \exp\br{\alpha / 2 c}}{\abs{\mathcal{R}_{\star_{u'}}} \exp\br{\star_{u'} \br{x} / 2 c}} \\
	~=~& \frac{\abs{\mathcal{R}}}{\abs{\mathcal{R}_{\star_{u'}}}} \exp\br{\frac{\alpha - \star_{u'} \br{x}}{2 c}}.
	\end{align*}
	The inequality follows from the observation that each $r \in \mathcal{R}$
	with $u'\br{x,r} \leq \alpha$ has un-normalized probability mass at most $\exp\br{\alpha / 2 c}$, and hence the entire set of such ``bad'' elements $r$ has total un-normalized probability mass at most $\abs{\mathcal{R}} \exp\br{\alpha / 2 c}$. In contrast, we know that there exist at least $\abs{\mathcal{R}_{\star_{u'}}} \geq 1$ elements with $u'\br{x,r} = \star_{u'} \br{x}$, and hence un-normalized probability mass $\abs{\mathcal{R}_{\star_{u'}}} \exp\br{\star_{u'} \br{x} / 2 c}$, and so this is a lower bound on the normalization term. The proof is completed by plugging in the appropriate value for $\alpha$, and by noting that 
	\begin{align*}
	    u \br{x,r} ~\leq~& u' \br{x,r} + \abs{u \br{x,r} - u' \br{x,r}} \\
	    ~\leq~& u' \br{x,r} + \max_{x,r} \abs{u \br{x,r} - u' \br{x,r}} .
	\end{align*}
\end{proof}

\subsection{Small Database Mechanism}
\label{app:smalldb}
\begin{reptheorem}{small-db-privacy-prop}
	\label{repthm:small-db-privacy-prop}
	If $\abs{q'_i - q'_j} \leq d_{\mathcal{X}}\br{i,j}, \forall{i,j \in \bs{N}} \text{ and } \\ \forall{q' \in \mathcal{Q}'}$, then the small database mechanism is $d_{\mathcal{X}}$-private. 
\end{reptheorem}
\begin{proof}
	First we will find the condition for $\Delta u' \br{i,j} \leq c d_{\mathcal{X}}\br{i,j}, \forall{i,j \in \bs{N}} \text{ and } \forall{q' \in \mathcal{Q}'}$: 
	\[
	\Delta u' \br{i,j} = \max_{y \in \mathcal{R}}{\max_{\substack{x,x' \in \mathbb{R}^N : \norm{x-x'}_1 \leq 2 ,\\ x_i \neq x'_i , x_j \neq x'_j \text{ for } i,j \in \bs{N}}}{\abs{u'\br{x,y} - u'\br{x',y}}}} .
	\]
	For some $x,x' \in \mathbb{N}^N$ such that $\norm{x-x'}_1 \leq 2 , x_i \neq x'_i , x_j \neq x'_j \text{ for some } i,j \in \bs{N}$, we have:
	\begin{align*}
	& \abs{u'\br{x,y} - u'\br{x',y}} \\
	~=~& \abs{c \max_{q' \in \mathcal{Q}'}{\abs{q'\br{x'} - q'\br{y}}} - c \max_{q' \in \mathcal{Q}'}{\abs{q'\br{x} - q'\br{y}}}} \\
	~\overset{(i)}{\leq}~& c \max_{q' \in \mathcal{Q}'} \abs{\bc{\abs{q'\br{x'} - q'\br{y}} - \abs{q'\br{x} - q'\br{y}}}} \\
	~\overset{(ii)}{\leq}~& c \max_{q' \in \mathcal{Q}'}\abs{q'\br{x} - q'\br{x'}} \\
	~=~& c \max_{q' \in \mathcal{Q}'} \abs{\ip{q'}{x-x'}} \\
	~\overset{(iii)}{\leq}~& c \max_{q' \in \mathcal{Q}'} \abs{q'_i - q'_j} ,
	\end{align*}
	where $(i)$ due to the fact that $\abs{\max_x {\abs{a\br{x}}} - \max_x {\abs{b\br{x}}}} \leq \max_x \abs{\bc{\abs{a\br{x}} - \abs{b\br{x}}}}$, $(ii)$ is by triangle inequality, and $(iii)$ is due to the choice of $x$ and $x'$. Thus we require 
	\[
	\Delta u' \br{i,j} \leq c \max_{q' \in \mathcal{Q}'} \abs{q'_i - q'_j} \leq c d_{\mathcal{X}} \br{i,j} .
	\]	
	The Small Database mechanism is simply an instantiation of the
	$\mathcal{M}_{\mathrm{Exp},c}\br{\cdot,u'}$ mechanism. Therefore, privacy follows from Theorem~\ref{exp-privacy-theorem}.
\end{proof}

We use the following theorem from \cite{dwork2014algorithmic} directly.  
\begin{theorem}[Theorem 4.2, \cite{dwork2014algorithmic}]
	\label{temp-smalldb-utility}
	For any finite class of linear queries $\mathcal{Q}'$, if $\mathcal{R} = \bc{y \in \mathbb{N}^N : \norm{y}_1 = \frac{\log{\abs{\mathcal{Q}'}}}{\alpha^2}}$ then for all $x \in \mathbb{N}^N$, there exists a $y \in \mathcal{R}$ such that:
	\[
	\max_{q' \in \mathcal{Q}'}{\abs{c q'\br{x} - c q'\br{y}}} \leq \alpha n .
	\]
\end{theorem}

\begin{repproposition}{utility-smalldb-prop}
	\label{repthm:utility-smalldb-prop}
	Let $\mathcal{Q}$ be any class of linear queries. Let $y$ be the
	database output by $\mathrm{SmallDB} \\ \br{x,\mathcal{Q}',c,\alpha}$. Then with probability $1-\beta$:
	\begin{align*}
	    \max_{q \in \mathcal{Q}}{\abs{q\br{x} - cq'\br{y}}}
	    \leq& n \max_{q \in \mathcal{Q}}\norm{q-cq'}_\infty + \alpha n \\
	    & + 2 c\bc{\frac{\log N \log \abs{\mathcal{Q}}}{\alpha^2} + \log\br{\frac{1}{\beta}}} .
	\end{align*}
\end{repproposition}
\begin{proof}
	Applying the utility bounds for the $\mathcal{M}_{\mathrm{Exp},c}\br{\cdot , u'}$ mechanism (Theorem~\ref{exp-utility-theorem}) with $-\star_{u'} \br{x} \geq \alpha n$ (which follows from Theorem~\ref{temp-smalldb-utility}), we find:
	\[
	\P{\max_{q' \in \mathcal{Q}'}{\abs{c q'\br{x} - c q'\br{y}}} \geq \alpha n + 2 c \bc{\ln\br{\abs{\mathcal{R}}} + t}} \leq e^{-t} . 
	\]
	By noting that $\mathcal{R}$, which is the set of all databases of size at most $\log{\abs{\mathcal{Q}}} / {\alpha^2}$ (since $\abs{\mathcal{Q}'} = \abs{\mathcal{Q}}$), satisfies $\abs{\mathcal{R}} \leq \abs{\mathcal{X}}^{\log{\abs{\mathcal{Q}}} / {\alpha^2}}$ and by setting $t=\log\br{\frac{1}{\beta}}$, we get with probability $1-\beta$:
	\begin{align*}
	& \max_{q' \in \mathcal{Q}'}{\abs{c q'\br{x} - c q'\br{y}}} \\
	~\leq~& \alpha n + 2 c \bc{\frac{\log N \log \abs{\mathcal{Q}}}{\alpha^2} + \log\br{\frac{1}{\beta}}} .
	\end{align*}
	Thus with probability $1-\beta$ we have ($q' \in \mathcal{Q}'$ is the one-to-one mapping of $q \in \mathcal{Q}$):
	\begin{align*}
	& \max_{q \in \mathcal{Q}}{\abs{q\br{x} - c q'\br{y}}} \\
	~\overset{(i)}{\leq}~& \max_{q \in \mathcal{Q}}\bc{\abs{q\br{x} - c q'\br{x}} + \abs{c q'\br{x} - c q'\br{y}}} \\
	~\overset{(ii)}{\leq}~& \max_{q \in \mathcal{Q}}\abs{q\br{x} - c q'\br{x}}  + \max_{q \in \mathcal{Q}}\abs{c q'\br{x} - c q'\br{y}} \\
	~=~& \max_{q \in \mathcal{Q}}\abs{\ip{q-c q'}{x}}  + \max_{q \in \mathcal{Q}}\abs{c q'\br{x} - c q'\br{y}} \\
	~\overset{(iii)}{\leq}~& \norm{x}_1 \max_{q \in \mathcal{Q}}\norm{q-c q'}_\infty  + \max_{q \in \mathcal{Q}}\abs{c q'\br{x} - c q'\br{y}} \\
	~\overset{(iv)}{=}~& n \max_{q \in \mathcal{Q}}\norm{q-c q'}_\infty  + \max_{q' \in \mathcal{Q}'}\abs{c q'\br{x} - c q'\br{y}} \\
	~\leq~& n \max_{q \in \mathcal{Q}}\norm{q-c q'}_\infty  + \alpha n \\
	& \quad + 2 c \bc{\frac{\log N \log \abs{\mathcal{Q}}}{\alpha^2} + \log\br{\frac{1}{\beta}}} ,
	\end{align*}
	where $(i)$ is by triangle inequality, $(ii)$ is by the fact that $\max_x \bc{a(x) + b(x)} \leq \max_x a(x) + \max_x b(x)$, $(iii)$ is by the H\"older's Inequality, and $(iv)$ is by the fact that $\norm{x}_1 = n$. 
\end{proof}

\begin{reptheorem}{small-db-utitlity-theorem}
	\label{repthm:small-db-utitlity-theorem}
	By the appropriate choice of $\alpha$, letting $y$ be the database
	output by $\mathrm{SmallDB}\br{x,\mathcal{Q}',c,\frac{\alpha}{2}}$, we can ensure that with probability $1-\beta$:
	\begin{align}
	    & \max_{q \in \mathcal{Q}}{\abs{q\br{x} - cq'\br{y}}} \nonumber \\
	    ~\leq~& n \max_{q \in \mathcal{Q}}\norm{q-cq'}_\infty + \br{cn^2 \gamma}^{1/3} , \label{small-db-utility-eq-1}
	\end{align}
	where $\gamma = 16 \log N \log \abs{\mathcal{Q}} + 4 \log\br{\frac{1}{\beta}}$. Equivalently, for any $c$ such that
	\begin{equation}
	\label{small-db-utility-eq-c-1}
	c ~\leq~ \frac{\alpha^3 n}{\gamma}
	\end{equation}
	with probability $1-\beta$: $\max_{q \in \mathcal{Q}}{\abs{q\br{x} - cq'\br{y}}} \leq n \max_{q \in \mathcal{Q}} \norm{q-cq'}_\infty  + \alpha n$.
\end{reptheorem}
\begin{proof}
	By Proposition~\ref{utility-smalldb-prop}, we get:
	\begin{align*}
	    & \max_{q \in \mathcal{Q}}{\abs{q\br{x} - c q'\br{y}}} \\ 
	    ~\leq~& n \max_{q \in \mathcal{Q}}\norm{q-cq'}_\infty  + \frac{\alpha}{2} n \\
	    & \quad + 2 c \bc{\frac{4 \log N \log \abs{\mathcal{Q}}}{\alpha^2} + \log\br{\frac{1}{\beta}}} .
	\end{align*}
	Setting this quantity to be at most $n \max_{q \in \mathcal{Q}}\norm{q-c q'}_p  + \alpha n$ and solving for $c$ yields \eqref{small-db-utility-eq-c-1}. Solving for $\alpha$ yields \eqref{small-db-utility-eq-1}.
\end{proof}

\subsection{Multiplicative Weights Exponential Mechanism}
\label{app:mwem}
\begin{reptheorem}{mwem-privacy-theorem}
	\label{repthm:mwem-privacy-theorem}
	If $\abs{q'_i - q'_j} \leq d_{\mathcal{X}}\br{i,j}, \forall{i,j \in \bs{N}} \text{ and } \\ \forall{q' \in \mathcal{Q}'}$, then the MWEM mechanism is $d_{\mathcal{X}}$-private.
\end{reptheorem}
\begin{proof}
	\textbf{Exponential Mechanism:} Consider the utility function $u': \mathbb{N}^N \times \mathcal{Q}' \rightarrow \mathbb{R}$ given by
	\[
	u' \br{x , q'} ~:=~ c \abs{q'\br{y} - q'\br{x}} , \text{ for some } y \in \mathbb{R}^N .
	\]
	First we find a condition for $\Delta u' \br{i,j} \leq c d_{\mathcal{X}}\br{i,j}, \forall{i,j \in \bs{N}} \text{ and } \forall{q' \in \mathcal{Q}'}$: 
	\begin{align*}
	& \Delta u' \br{i,j} \\
	~=~& \max_{q' \in \mathcal{Q}'}{\max_{\substack{x,x' \in \mathbb{N}^N : \norm{x-x'}_1 \leq 2 ,\\ x_i \neq x'_i , x_j \neq x'_j \text{ for } i,j \in \bs{N}}}{\abs{u'\br{x,q'} - u'\br{x',q'}}}} .
	\end{align*}
	For some $x,x' \in \mathbb{N}^N$ such that $\norm{x-x'}_1 \leq 2 , x_i \neq x'_i , x_j \neq x'_j \text{ for some } i,j \in \bs{N}$, we have:
	\begin{align*}
	& \abs{u'\br{x,q'} - u'\br{x',q'}} \\
	~=~& \abs{c \abs{q'\br{y} - q'\br{x}} - c \abs{q'\br{y} - q'\br{x'}}} \\
	~\overset{(i)}{\leq}~& c \abs{q'\br{x'} - q'\br{x}} \\
	~=~& c \abs{\ip{q'}{x-x'}} \\
	~\overset{(ii)}{\leq}~& c \abs{q'_i - q'_j} ,
	\end{align*}
	where $(i)$ is by triangle inequality, and $(ii)$ is due to the choice of $x$ and $x'$. That is we require 
	\[
	\Delta u' \br{i,j} ~\leq~ c \abs{q'_i - q'_j} ~\leq~ c d_{\mathcal{X}} \br{i,j} .
	\]	
	Thus with the above transformed class $\mathcal{Q}'$, if we use the $\mathcal{M}_{\mathrm{Exp},2 c T}\br{x,u'}$ mechanism, we get $\frac{d_{\mathcal{X}} \br{i,j}}{2 T}$-privacy. 
	
	\textbf{Laplace Mechanism:} If $\abs{q'_i - q'_j} \leq d_{\mathcal{X}}\br{i,j}, \forall{i,j \in \bs{N}} \text{ and } \forall{q' \in \mathcal{Q}'}$, then the Laplace mechanism given by $m = c q'\br{x} + \mathrm{Lap}\br{2 c T}$ satisfies $\frac{d_{\mathcal{X}} \br{i,j}}{2 T}$-privacy.
	
	The composition rules for $d_{\mathcal{X}}$-privacy state that $c$ values accumulate appropriately. We make $T$ calls to the Exponential Mechanism with parameter $2cT$ and $T$ calls to the Laplace
	Mechanism with parameter $2cT$, resulting in $d_{\mathcal{X}}$-privacy.
\end{proof}

\begin{reptheorem}{mwem-utility-theorem}
	\label{repthm:mwem-utility-theorem}
	For any dataset $x$, set of linear queries $\mathcal{Q}$, $T \in \mathbb{N}$, and $c> 0$, with probability at least $1 - 2 T / \abs{\mathcal{Q}}$, MWEM produces $y$ such that 
	\begin{align*}
	& \max_{q \in \mathcal{Q}}\abs{cq'\br{y} - q\br{x}} \\
	~\leq~& 2 n \sqrt{\frac{\log N}{T}} + 10 T c\log\abs{\mathcal{Q}} + n \max_{q \in \mathcal{Q}} \norm{cq'-q}_\infty .
	\end{align*}
	By setting $2 n \sqrt{\frac{\log N}{T}} = 10 T c\log\abs{\mathcal{Q}}$, we get
	\begin{align*}
	& \max_{q \in \mathcal{Q}}\abs{cq'\br{y} - q\br{x}} \\
	~\leq~& n \max_{q \in \mathcal{Q}} \norm{cq'-q}_\infty + \frac{20}{5^{2/3}} \br{n^2 \log N \log\abs{\mathcal{Q}}}^{1/3} c^{1/3} .
	\end{align*}
\end{reptheorem}
\begin{proof}
	The following inequality follows directly by replacing the $\epsilon$ by $\frac{1}{c}$ along the proof given in \cite{hardt2012simple}: 
	\[
	\max_{q' \in \mathcal{Q}'}\abs{c q'\br{y} - c q'\br{x}} \leq 2 n \sqrt{\frac{\log N}{T}} + 10 T c \log\abs{\mathcal{Q}} .
	\]
	Then with probability at least $1 - 2 T / \abs{\mathcal{Q}}$, we have
	\begin{align*}
	& \max_{q \in \mathcal{Q}}\abs{c q'\br{y} - q\br{x}} \\
	~\leq~& \max_{q \in \mathcal{Q}} \bc{\abs{c q'\br{y} - c q'\br{x}} + \abs{c q'\br{x} - q\br{x}}} \\
	~\leq~& \max_{q' \in \mathcal{Q}'} \abs{c q'\br{y} - c q'\br{x}} + \max_{q \in \mathcal{Q}} \abs{c q'\br{x} - q\br{x}} \\
	~=~& \max_{q' \in \mathcal{Q}'} \abs{c q'\br{y} - c q'\br{x}} + \max_{q \in \mathcal{Q}} \abs{\ip{c q'-q}{x}} \\
	~\leq~& \max_{q' \in \mathcal{Q}'} \abs{c q'\br{y} - c q'\br{x}} + \max_{q \in \mathcal{Q}} \norm{c q'-q}_\infty\norm{x}_1 \\
	~\leq~& 2 n \sqrt{\frac{\log N}{T}} + 10 T c \log\abs{\mathcal{Q}} + n \max_{q \in \mathcal{Q}} \norm{c q'-q}_\infty .
	\end{align*} 
\end{proof}

\section{Pre-processing Optimization}
\label{sec:biopt}

In Section~\ref{sec:mech-linear-query}, we have shown that by (approximately) solving certain pre-processing optimization problems (\emph{e.g.} \eqref{opt-laplace-multi-linear-query},\eqref{opt-smalldb}), we can obtain the model parameters of the $d_{\mathcal{X}}$-private mechanisms with enhanced utility. One can easily verify that these problems are non-convex optimization problems. Recently, in the optimization and machine learning community, there is a huge interest in developing efficient algorithms for non-convex optimization problems with provable guarantees. One can also observe that these pre-processing optimization problems exhibit coordinate friendly structures, and thus the coordinate descent family of algorithms \cite{wright2015coordinate} is a natural choice to solve them. 

Consider the optimization problem \eqref{opt-laplace-multi-linear-query} under squared loss. One can easily verify that  $f\br{c,Q'} = n^2 \norm{c\odot Q' - Q}_{2}^2 + 2 \norm{c}_2^2$ is a (smooth) multi-convex function \emph{i.e.} $f\br{\cdot,Q'}$ is convex in $c$ for any fixed $Q'$, and $f\br{c,\cdot}$ is convex in $Q'$ for any fixed $c$, but $f$ is not jointly convex in $\br{c,Q'}$. Recently \cite{xu2013block} have shown that under certain conditions, the multi-convex optimization problem can be efficiently solved via a variant of cyclic block coordinate descent algorithm. Now consider the optimization problem \eqref{opt-smalldb} with the objective function $f\br{c,\mathcal{Q}'} = n \max_{q \in \mathcal{Q}} \norm{cq'-q}_\infty + \frac{20}{5^{2/3}} \br{n^2 \log N \log\abs{\mathcal{Q}}}^{1/3} c^{1/3}$. In this case, the objective function is both non-smooth and non-convex, thus the resulting problem is very hard to optimize. However, in practice approximate solutions would still yield good utility.


\section{Statistical Queries}
\label{sec:stat-query}
A statistical query on a data universe $\mathcal{X} \subset \mathbb{R}^d$ is defined by a mapping $q: \mathcal{X} \rightarrow \mathcal{Y} \subset \mathbb{R}^k$. Abusing notation, we define the evaluation of a statistical query $q$ on the database $x \in \mathcal{X}^n$ to be the average of the predicate over the rows
\begin{equation}
\label{count-query}
q\br{x} = \frac{1}{n} \sum_{i=1}^{n}{q\br{x_i}} .
\end{equation}
When $q\br{u} = u$, $\forall u \in \mathcal{X}$, we call it $d$-way marginal query. We can actually treat the statistical query as a linear query over histogram ($y \in \mathbb{N}^N$) with query matrix $Q \in \mathbb{R}^{k \times N}$. But we can exploit the sum-structure (\eqref{count-query}) of it to design efficient algorithms.  

\begin{definition}
	\label{generalized-global-sensitivity-def}
	For $u,v \in \mathcal{X}$ (with $u \neq v$), define the generalized global sensitivity of a query $q \in \mathcal{Q}$ (w.r.t. $\norm{\cdot}$) as
	\[
	\Delta_{\norm{\cdot}}^q \br{u,v} ~:=~ \max_{\substack{x,x' \in \mathcal{X}^n : \norm{x-x'}_H \leq 1 ,\\ x_i = u , x'_i = v \text{ for } i \in \bs{n}}}{\norm{q\br{x} - q\br{x'}}} .
	\]
	Also define $\Delta_{\norm{\cdot}}^q := \max_{u,v \in \mathcal{X}}{\Delta_{\norm{\cdot}}^q \br{u,v}}$ (the usual global sensitivity). When $\norm{\cdot} = \norm{\cdot}_p$, we simply write $\Delta_p^q$.
\end{definition}
The generalized global sensitivity (for $u,v \in \mathcal{X}$) of the statistical query $q$ is given by 
\begin{align*}
& \max_{\substack{x,x' \in \mathcal{X}^n : \norm{x-x'}_H \leq 1 ,\\ x_i = u, x'_i = v \text{ for } i \in \bs{n}}}{\norm{\frac{1}{n} \sum_{i=1}^{n}{q\br{x_i}} - \frac{1}{n} \sum_{i=1}^{n}{q\br{x'_i}}}} \\
~=~& \frac{\norm{q\br{u} - q\br{v}}}{n} . 
\end{align*}
For the $d$-way marginal query $q$, we have $\Delta_{\norm{\cdot}}^q \br{u,v} = \frac{\norm{u - v}}{n}$.

\begin{definition}
	\label{dx-privacy-def}
	Let $\mathcal{X}$ (with $\phi \in \mathcal{X}$) be the data universe, $d_{\mathcal{X}}: \mathcal{X} \times \mathcal{X} \rightarrow \mathbb{R}$ be the privacy budget, and $q: \mathcal{X}^n \rightarrow \mathcal{Y}$ be the query. A mechanism $\mathcal{M}: \mathcal{X}^n \times \mathcal{Q} \rightsquigarrow \mathcal{Y}$ is said to be $d_{\mathcal{X}}$-private iff $\forall{x,x' \in \mathcal{X}^n}$ s.t. $\norm{x-x'}_H \leq 1$, and $x_i \neq x'_i$ (for some $i \in \bs{n}$), $\forall S \subseteq \mathcal{Y}$ and $\forall{q \in Q}$ we have
	\[
	\frac{\P{\mathcal{M}\br{x,q} \in S}}{\P{\mathcal{M}\br{x',q} \in S}} ~\leq~ \exp\br{d_{\mathcal{X}} \br{x_i , x'_i}} .
	\]
	When $d_{\mathcal{X}} \br{u , v} = \epsilon, \forall{u,v \in \mathcal{X}}$, we recover the standard $\epsilon$-differential privacy, and when $d_{\mathcal{X}} \br{u,v} = \epsilon_u \wedge \epsilon_v$ for $u,v \in \mathcal{X}$, we recover the instance specific differential privacy notion introduced in \cite{ghosh2015selling}. 
\end{definition} 

For a given query $q : \mathcal{X}^n \rightarrow \mathcal{Y} \subset \mathbb{R}^k$ over the database $x \in \mathcal{X}^n$, consider the following variant of Laplace mechanism (with the mapping $\mathcal{X} \mapsto \mathcal{X}'$, and $c \in \mathbb{R}^k$): 
\begin{equation}
\label{weight-laplace-stat}
\textsf{Z} ~=~ \mathcal{M}_{\mathrm{Lap},c}\br{x' , q} ~:=~ c \odot q\br{x'} + \br{\textsf{Y}_1,\dots ,\textsf{Y}_k} ,
\end{equation}
where $\textsf{Y}_i \overset{\perp}{\sim} \mathrm{Lap}\br{c_i}$. Below we show that the above variant of Laplace mechanism satisfies the $d_{\mathcal{X}}$-privacy under a sensitivity bound condition. 
\begin{theorem} 
	\label{laplace-privacy-prop-stat}
	Let $q'\br{u} := q\br{u'}, \forall{u \in \mathcal{X}}$ under the mapping $\mathcal{X} \mapsto \mathcal{X}'$. If $\Delta_1^{q'} \br{u,v} \leq d_{\mathcal{X}}\br{u,v}$, $\forall{u,v \in \mathcal{X}}$, then the mechanism $\mathcal{M}_{\mathrm{Lap},c} \br{x' , q}$ given by \eqref{weight-laplace-stat} satisfies the $d_{\mathcal{X}}$-privacy. 
\end{theorem} 
\begin{proof}
	Proof is similar to that of Theorem~\ref{laplace-privacy-prop}.
\end{proof}
The sensitivity bound condition of the above theorem for a statistical query $q$ can be written as follows:
\[
\Delta_1^{q'} \br{u,v} ~=~ \frac{\norm{q\br{u'} - q\br{v'}}_1}{n} ~\leq~ d_{\mathcal{X}}\br{u,v} , \quad \forall{u,v \in \mathcal{X}} .
\]
For the $d$-way marginal query, the above condition reduces to $\norm{u' - v'}_1 \leq n d_{\mathcal{X}}\br{u,v} , \forall{u,v \in \mathcal{X}}$. The next theorem characterizes the performance of the $\mathcal{M}_{\mathrm{Lap},c}\br{x' , q}$ mechanism under different choices of utility measures:
\begin{theorem}
	\label{laplace-utility-prop-stat}
	Let $q: \mathcal{X}^n \rightarrow \mathbb{R}^k$ be a statistical query of the form $q\br{x} = \frac{1}{n} \sum_{i=1}^{n}{q\br{x_i}}$, and let $\textnormal{\textsf{Z}} = \mathcal{M}_{\mathrm{Lap},c}\br{x' , q} = c \odot q\br{x'} + \textnormal{\textsf{Y}}$ with $\textnormal{\textsf{Y}}_i \overset{\perp}{\sim} \mathrm{Lap}\br{c_i}$. 
	\begin{enumerate}
		\item When $\ell_2^2 \br{y,y'} = \norm{y - y'}_2^2$, we have
		\begin{align*}
		& \mathrm{err}_{\ell_2^2}\br{\mathcal{M}_{\mathrm{Lap},c},q} \\
		~\leq~& 2 \bc{\max_{u \in \mathcal{X}} \norm{c \odot q\br{u'} - q\br{u}}_2^2 + 2 \norm{c}_2^2} .
		\end{align*}
		\item When $\ell_p \br{y,y'} = \norm{y - y'}_p$, we have
		\begin{align*}
		& \mathrm{err}_{\ell_p}\br{\mathcal{M}_{\mathrm{Lap},c},q} \\
		~\leq~& \max_{u \in \mathcal{X}} \norm{c \odot q\br{u'} - q\br{u}}_p + \Ee{Y_i \overset{\perp}{\sim} \mathrm{Lap}\br{c_i}}{\norm{\textnormal{\textsf{Y}}}_p} .
		\end{align*}
		\item $\forall{\delta \in (0,1]}$, with probability at least $1-\delta$ we have 
		\begin{align*}
		& \norm{\textnormal{\textsf{Z}} - q\br{x}}_\infty  \\
		~\leq~& \max_{u \in \mathcal{X}} \norm{c \odot q\br{u'} - q\br{u}}_{\infty} + \ln\br{\frac{k}{\delta}} \cdot \norm{c}_{\infty} .
		\end{align*}
	\end{enumerate}
\end{theorem}
\begin{proof}
	Proof is similar to that of Theorem~\ref{laplace-utility-prop}, but with the following change in the appropriate places (with $q'\br{u} := q\br{u'}, \forall{u \in \mathcal{X}}$):
	\begin{align*}
	& \max_{x \in \mathcal{X}^n} \norm{c \odot q'\br{x} - q\br{x}}_p \\
	~=~& \max_{x \in \mathcal{X}^n} \frac{1}{n} \norm{c \odot \sum_{i=1}^{n}{q\br{x'_i}} - \sum_{i=1}^{n}{q\br{x_i}}}_p \\
	~\leq~& \max_{x \in \mathcal{X}^n} \frac{1}{n} \sum_{i=1}^{n}{\norm{c \odot q\br{x'_i} - q\br{x_i}}_p} \\
	~=~& \frac{1}{n} \sum_{i=1}^{n}{\max_{x_i \in \mathcal{X}} \norm{c \odot q\br{x'_i} - q\br{x_i}}_p} \\ 
	~=~& \max_{u \in \mathcal{X}} \norm{c \odot q\br{u'} - q\br{u}}_p .
	\end{align*}
\end{proof}

Now we model the following optimization problem to select the model parameters $c$ and $\mathcal{X} \mapsto \mathcal{X}'$ of the $\mathcal{M}_{\mathrm{Lap},c}\br{x',q}$ mechanism:
\begin{equation}
\label{opt-laplace-stat-query}
\begin{aligned}
& \underset{c , \mathcal{X}'}{\text{minimize}}
& & f_{\ell,\mathcal{M}}\br{c,\mathcal{X}';q,n} \\
& \text{subject to}
& & \norm{q\br{u'} - q\br{v'}}_1 ~\leq~ n d_{\mathcal{X}}\br{u,v} ,  \forall{u,v \in \mathcal{X}} \\
&&& c \succeq 0.
\end{aligned}
\end{equation}
The objective function $f_{\ell,\mathcal{M}}\br{c,\mathcal{X}';q,n}$ depends on the utility function that we are interested in. For example, when $\ell_2^2 \br{y,y'} = \norm{y - y'}_2^2$, we can choose $f_{\ell_2^2,\mathcal{M}}\br{c,\mathcal{X}';q,n} = \max_{u \in \mathcal{X}} \norm{c \odot q\br{u'} - q\br{u}}_2^2 + 2 \norm{c}_2^2$. In fact there are two ways to design $d_{\mathcal{X}}$-private mechanisms from existing $\epsilon$ differentially private mechanisms: either transform the query vector or the data universe. The approach we used above is $\mathcal{X} \mapsto \mathcal{X}'$ (that is $q\br{u} \rightarrow q'\br{u} = q\br{u'}$). Thus we can reduce the number of variables in the pre-processing optimization by a factor of $k$.

Consider a privacy budget (metric) of the form $d_{\mathcal{X}}\br{u,v} = \sum_{i=1}^{d}{d_i \br{u_i , v_i}}$, where for example $d_i \br{u_i , v_i} = \epsilon_i \ind{u_i \neq v_i}$. In this case, if $\abs{u'_i - v'_i} \leq n d_i \br{u_i , v_i} , \forall{u_i,v_i \in \mathcal{X}_i}$ (for example, when $\mathcal{X} = \bc{-1,+1}^d$, we have $\mathcal{X}_i = \bc{-1,+1}$), then the $d$-way marginal query is $d_{\mathcal{X}}$-private. Moreover, when $\ell_1 \br{y,y'} = \norm{y - y'}_1$, we have 
\[
f_{\ell_1,\mathcal{M}}\br{c,\mathcal{X}';q,n} ~=~  \sum_{i=1}^{k}{f_i\br{c_i,\mathcal{X}_i}}
\]
with $f_i\br{c_i,\mathcal{X}_i} = \max_{u_i \in \mathcal{X}_i} \abs{c_i u'_i - u_i} + c_i$ for $d$-way marginal queries (since $\max_{u \in \mathcal{X}} \norm{c \odot q\br{u'} - q\br{u}}_1 + \Ee{\textsf{Y}_i \overset{\perp}{\sim} \mathrm{Lap}\br{c_i}}{\norm{\textsf{Y}}_1} = \max_{u \in \mathcal{X}} \sum_{i=1}^{k}{\abs{c_i u'_i - u_i}} + \norm{c}_1 = \sum_{i=1}^{k}{\max_{u_i \in \mathcal{X}_i} \abs{c_i u'_i - u_i}} + \sum_{i=1}^{k}{c_i}$). Thus in this setting, we can instantiate and \textit{relax} the above optimization problem \eqref{opt-laplace-stat-query} into $k$ independent optimization problems as follows:
\begin{equation}
\label{opt-marginal-query-1}
\begin{aligned}
& \underset{c_i , {\mathcal{X}}'_i}{\text{minimize}}
& & \max_{u_i \in \mathcal{X}_i} \abs{c_i u'_i - u_i} + c_i \\
& \text{subject to}
& & \abs{u'_i - v'_i} ~\leq~ n d_i \br{u_i , v_i} , \quad \forall{u_i,v_i \in \mathcal{X}_i} \\
&&& c_i \geq 0.
\end{aligned}
\end{equation}


\newpage 
\begin{thebibliography}{99}
\bibitem{alaggan2015heterogeneous} M.~Alaggan, S.~Gambs, and A.~M.~Kermarrec. Heterogeneous differential privacy. \textit{arXiv preprint arXiv:1504.06998}, 2015.
\bibitem{andres2013geo} M.~E.~Andr{\'e}s, N.~E.~Bordenabe, K.~Chatzikokolakis, and C.~Palamidessi. Geo-indistinguishability: Differential privacy for location-based systems. \textit{Proceedings of the ACM SIGSAC conference on Computer \& communications security}, pages 901--914, 2013.
\bibitem{barak2007privacy} B.~Barak, K.~Chaudhuri, C.~Dwork, S.~Kale, F.~McSherry, and K.~Talwar. Privacy, accuracy, and consistency too: a holistic solution to contingency table release. \textit{Proceedings of the ACM SIGMOD-SIGACT-SIGART symposium on Principles of database systems}, pages 273--282, 2007.
\bibitem{blum2005practical} A.~Blum, C.~Dwork, F.~McSherry, and K.~Nissim. Practical privacy: the SuLQ framework. \textit{Proceedings of the ACM SIGMOD-SIGACT-SIGART symposium on Principles of database systems}, pages 128--138, 2005.
\bibitem{blum2013learning} A.~Blum, K.~Ligett, and A.~Roth. A learning theory approach to noninteractive database privacy. \textit{Journal of the ACM (JACM)}, 60(2):12, 2013.
\bibitem{chatzikokolakis2013broadening} K.~Chatzikokolakis, M.~E.~Andr{\'e}s, N.~E.~Bordenabe, and C.~Palamidessi. Broadening the scope of differential privacy using metrics. \textit{International Symposium on Privacy Enhancing Technologies Symposium}, pages 82--102, 2013.
\bibitem{dwork2012fairness} C.~Dwork, M.~Hardt, T.~Pitassi, O.~Reingold, and R.~Zemel. Fairness through awareness. \textit{Proceedings of the Innovations in Theoretical Computer Science Conference}, pages 214--226, 2012.
\bibitem{dwork2006calibrating} C.~Dwork, F.~McSherry, K.~Nissim, and A.~Smith. Calibrating Noise to Sensitivity in Private Data Analysis. \textit{Proceedings of the Conference on Theory of Cryptography}, pages 265--284, 2006.
\bibitem{dwork2014algorithmic} C.~Dwork, and A.~Roth. The algorithmic foundations of differential privacy. \textit{Foundations and Trends{\textregistered} in Theoretical Computer Science}, pages 211--407, 2014.
\bibitem{fienberg2010differential} S.~E.~Fienberg, A.~Rinaldo, and X.~Yang. Differential privacy and the risk-utility tradeoff for multi-dimensional contingency tables. In \textit{International Conference on Privacy in Statistical Databases}, pages 187--199, 2010.
\bibitem{ghosh2015selling} A.~Ghosh, and A.~Roth. Selling privacy at auction. \textit{Games and Economic Behavior}, pages 334--346, 2015.
\bibitem{ghosh2012universally} A.~Ghosh, T.~Roughgarden, and M.~Sundararajan. Universally utility-maximizing privacy mechanisms. \textit{SIAM Journal on Computing}, pages 1673--1693, 2012.
\bibitem{haney-blowfish} S.~Haney, A.~Machanavajjhala, and B.~Ding. Design of policy-aware differentially private algorithms. \textit{Proceedings of the VLDB Endowment}, pages 264--275, 2015.
\bibitem{hardt2012simple} M.~Hardt, K.~Ligett, and F.~McSherry. A simple and practical algorithm for differentially private data release. \textit{Advances in Neural Information Processing Systems}, pages 2339--2347, 2012.
\bibitem{hardt2010geometry} M.~Hardt, and K.~Talwar. On the geometry of differential privacy. \textit{Proceedings of the ACM symposium on Theory of computing}, pages 705--714, 2010.
\bibitem{he2014blowfish} X.~He, A.~Machanavajjhala, and B.~Ding. Blowfish privacy: Tuning privacy-utility trade-offs using policies. \textit{Proceedings of the ACM SIGMOD international conference on Management of data}, pages 1447--1458, 2014.
\bibitem{jorgensen2015conservative} Z.~Jorgensen, T.~Yu, and G.~Cormode. Conservative or liberal? Personalized differential privacy. \textit{International Conference on Data Engineering}, pages 1023--1034, 2015.
\bibitem{li2011efficient} C.~Li, and G.~Miklau. Efficient batch query answering under differential privacy. \textit{arXiv preprint arXiv:1103.1367}, 2011.
\bibitem{li2012adaptive} C.~Li, and G.~Miklau. An adaptive mechanism for accurate query answering under differential privacy. \textit{Proceedings of the VLDB Endowment}, pages 514--525, 2012.
\bibitem{elevation-API} Google Maps. Google Elevation API. \url{https://developers.google.com/maps/documentation/elevation/intro}, 2018.
\bibitem{mcsherry2007mechanism} F.~McSherry, and K.~Talwar. Mechanism design via differential privacy. In \textit{Foundations of Computer Science}, pages 94--103, 2007.
\bibitem{salil-tut} V.~Vadhan. The complexity of Differential Privacy. \textit{Tutorials on the Foundations of Cryptography}, pages 347--450, Springer, 2017.
\bibitem{USA-cities} SimpleMaps. United States Cities Database. \url{https://simplemaps.com/data/us-cities}, 2018.
\bibitem{wright2015coordinate} S.~J.~Wright. Coordinate descent algorithms. \textit{Mathematical Programming}, pages 3--34, 2015.
\bibitem{xu2013block} Y.~Xu, and W.~Yin. A block coordinate descent method for regularized multiconvex optimization with applications to nonnegative tensor factorization and completion. \textit{SIAM Journal on imaging sciences}, pages 1758--1789, 2013.
\end{thebibliography}
\end{document}